\documentclass{article}

\usepackage[preprint]{neurips_2025}
\usepackage[hyphens]{url}

\usepackage[utf8]{inputenc} %
\usepackage[T1]{fontenc}    %
\usepackage{hyperref}       %
\usepackage{url}            %
\usepackage{booktabs}       %
\usepackage{amsfonts}       %
\usepackage{graphicx}
\usepackage{amsmath}
\usepackage{cleveref}
\usepackage{nicefrac}       %
\usepackage{microtype}      %
\usepackage{xcolor}         %
\usepackage{natbib}
\usepackage{colortbl}
\usepackage{xcolor}
\usepackage{multirow}
\usepackage{tcolorbox}
\usepackage{caption}
\usepackage{newfloat}
\usepackage{amssymb}
\usepackage{pifont}
\usepackage{subfigure}
\usepackage{CJKutf8}
\usepackage{enumitem}
\usepackage{xspace}
\usepackage{wrapfig}
\usepackage{arydshln}
\definecolor{best}{RGB}{255, 200, 0}   %
\definecolor{lightgreen}{HTML}{A8E6A9}
\definecolor{lightred}{HTML}{FFCCC9}
\definecolor{high}{RGB}{173, 216, 230}
\definecolor{green}{HTML}{228B22}
\definecolor{teal}{HTML}{008080}
\newcommand{\greenscriptsize}[1]{{\color{green}\scriptsize #1}}
\newcommand{\redscriptsize}[1]{{\color{red}\scriptsize #1}}

\newcommand{\RQone}{\texttt{RQ1}\xspace}
\newcommand{\RQtwo}{\texttt{RQ2}\xspace}
\newcommand{\RQthree}{\texttt{RQ3}\xspace}
\newcommand{\RQfour}{\texttt{RQ4}\xspace}

\usepackage{amsmath,stmaryrd,graphicx}

\makeatletter
\newcommand{\fixed@sra}{$\vrule height 2\fontdimen22\textfont2 width 0pt\shortrightarrow$}
\newcommand{\shortarrow}[1]{%
  \mathrel{\text{\rotatebox[origin=c]{\numexpr#1*45}{\fixed@sra}}}
}
\makeatother

\newtcolorbox[auto counter]{example}[2][]{
  width=\linewidth,
  sharp corners=all,
  colback=white!96!black,
  float,
  floatplacement=t,
  label=#2,
  title={Box \thetcbcounter. #1},
}

\title{Crosslingual Reasoning through Test-Time Scaling}

\author{%
\vspace{1mm} 
Zheng-Xin Yong$^{1\thanks{Core contributors. See Appendix~\ref{app:author-contributions} for further details.}}$\quad 
M. Farid Adilazuarda$^{2^*}$ \quad 
Jonibek Mansurov$^{2^*}$\quad 
Ruochen Zhang$^{1^*}$ 
\\
\vspace{1mm}
\textbf{Niklas Muennighoff}$^{3}$ \quad
\textbf{Carsten Eickhoff}$^4$ \quad
\textbf{Genta Indra Winata}$^5$ \quad
\textbf{Julia Kreutzer}$^6$ 
\\ 
\vspace{1.5mm}
\textbf{Stephen H.~Bach}$^1$ \quad \textbf{Alham Fikri Aji}$^2$\\
\vspace{1mm}
$^1$Brown University \quad $^2$MBZUAI \quad $^3$Stanford University \quad $^4$University of Tübingen \\
\vspace{1.5mm}
$^5$Capital One \quad $^6$Cohere Labs\\
\texttt{contact.yong@brown.edu}
}

\begin{document}

\maketitle

\begin{abstract}
  Reasoning capabilities of large language models are primarily studied for English, even when pretrained models are multilingual. 
  In this work, we investigate to what extent English reasoning finetuning with long chain-of-thoughts (CoTs) can generalize across languages. 
  First, we find that scaling up inference compute for English-centric reasoning language models (RLMs) improves multilingual mathematical reasoning across many languages including low-resource languages, to an extent where they outperform models \textit{twice their size}.
  Second, we reveal that while English-centric RLM's CoTs are naturally predominantly English, they consistently follow a \textit{quote-and-think} pattern to reason about quoted non-English inputs.
  Third, we discover an effective strategy to control the language of long CoT reasoning, and we observe that models reason better and more efficiently in high-resource languages. 
  Finally, we observe poor out-of-domain reasoning generalization, in particular from STEM to cultural commonsense knowledge, even for English. 
  Overall, we demonstrate the potentials, study the mechanisms and outline the limitations of crosslingual generalization of English reasoning test-time scaling. We conclude that practitioners should let English-centric RLMs reason in high-resource languages, while further work is needed to improve reasoning in low-resource languages and out-of-domain contexts.
  
\end{abstract}

\section{Introduction}

Scaling up compute at test-time can maximize model performance and output quality~\cite{snell2024scaling,brown2024large,wu2024inference,levi2024simple}, but it has been understudied in multilingual settings.
In particular, reasoning language models (RLMs), such as Deepseek's r1 \cite{guo2025deepseek} and OpenAI's o1 or o3 models \citep{jaech2024openaio1, openai2025o3}, strongly benefit from added inference compute to their long chain-of-thoughts (long CoTs) \citep{chen2025towards}. However, this advantage has primarily been explored in English contexts, such as in
recent work that combined small-scale reasoning finetuning with scaled up thinking tokens at test time~\citep{muennighoff2025s1,ye2025limo}. 
State-of-the-art RLMs rely on reasoning training data that contain long CoTs, which is currently most available for English~\citep{ghosh2025multilingualmindsurvey}. Thus, these RLMs are English-centric~\citep{muennighoff2025s1,hou2025advancing,hao2024training,gu2024critic,xiang2025towards,ghosh2025multilingualmindsurvey}. 
Given that their base models are often multilingual models such as Qwen models \citep{yang2024qwen25}, does reasoning finetuning in English give them multilingual reasoning abilities?

In this work, we investigate \emph{how much test-time compute can improve multilingual reasoning abilities of English-centric RLMs}. 
In particular, our research questions are as follows: 
\begin{enumerate}
    \item[\RQone.] 
    \textbf{Crosslingual test-time scaling}: How effective is test-time scaling of English-centric RLMs on multilingual reasoning tasks? (\Cref{sec:crosslingual-mgsm})
    \item[\RQtwo.] \textbf{Language-mixing behaviors}: What kind of language-mixing patterns do English-centric RLMs exhibit when they interact with non-English prompts? (\Cref{sec:language-mixing-behaviors})
    \item[\RQthree.] \textbf{Language forcing}: How well do English-centric RLMs perform when being forced to think in non-English languages? (\Cref{sec:language-forcing})
    \item[\RQfour.] \textbf{Cross-domain generalization:} How well does crosslingual reasoning generalize beyond the original STEM domain, such as humanities and social sciences? (\Cref{sec:cross-domain-generalization})
\end{enumerate}

 We experiment with s1 models \cite{muennighoff2025s1} as our English-centric RLMs for crosslingual generalization study. They are multilingual Qwen2.5-Instruct models \citep{yang2024qwen25} supervised finetuned on 1k training samples of English STEM reasoning tasks and achieve state-of-the-art performance on English math reasoning benchmarks \citep{muennighoff2025s1}. Our most significant contributions are as follows: \footnote{We release our code and artifacts at \url{https://github.com/BatsResearch/crosslingual-test-time-scaling}.}
\begin{enumerate}[leftmargin=*]
    \item We provide evidence that larger models benefit from crosslingual test-time scaling, which \textit{contrasts} with contemporary work \cite{son2025linguistic} that reports early negative findings based on 1.5B models. Crosslingual test-time scaling is not only effective for both high-resource and low-resource languages, but it also allows an RLM to outperform models twice its size on multilingual math reasoning tasks.
    \item We report a dominant language-mixing pattern where RLMs quote non-English phrases related to the question prompts in quotation marks in the thinking process. This \textit{quote-and-think} pattern suggests that model's multilingual capability to parse and understand questions enables crosslingual generalization of English reasoning finetuning.
    \item We discover an effective strategy to control the reasoning language of RLMs, and we find that forcing RLMs to think in high-resource languages yields substantially better reasoning performance  than in low-resource languages. Furthermore, the long CoTs for high-resource languages are more token-efficient at test time.
    \item We show that while reasoning finetuning may benefit tasks in certain domains, we \textit{do not} observe consistent benefits of scaling up thinking tokens for non-STEM tasks that are outside of the STEM domain of reasoning finetuning data. In domains requiring cultural commonsense knowledge, test-time scaling can even hurt performance.
\end{enumerate}

Our work shows that test-time scaling of English-centric RLMs can serve as a strong multilingual reasoning baseline. 
Furthermore, for English-centric reasoning finetuning, it is advisable to finetune models through data-efficient training (such as with s1's data \citep{muennighoff2025s1}) to minimize catastrophic forgetting of multilingual capability. We recommend letting the English-centric RLMs reason in high-resource languages such as English and Chinese for optimal performance and inference-compute efficiency. Future work is needed for enabling RLMs to generalize to out-of-domain tasks or to reason in low-resource languages.

\section{Background and Related Work}

\paragraph{Reasoning language models (RLMs)} Recent advancements of reasoning language models (RLMs) such as OpenAI-o1 \cite{jaech2024openaio1,openai2025o3} and DeepSeek-R1 \citep{guo2025deepseek} builds on LLMs' capability to perform intermediate reasoning steps, which is commonly referred to as chain-of-thought reasoning \citep{wei2022cot}. Prior work demonstrates that these intermediate computation steps can significantly improve the correctness for final answer outputs \citep{wei2022cot,ling-etal-2017-program,cobbe2021training,nye2021show,li2024chain}. Furthermore, extending the lengths of these computation steps, thereby creating long chain-of-thoughts (long CoTs), can allow the model to backtrack on incorrect reasoning steps and self-correct its final answer \citep{chen2025towards,gandhi2025cognitive,guo2025deepseek,hou2025advancing,lee2025evolving}. 
In our work, we focus on RLMs with long CoTs capability, which is an emerging research area.
These models are created through  \textit{distilling} long English-only reasoning chains from larger RLMs \citep{huang2024o1replication,ye2025limo,bespokelabs2025,slam-distillation-from-r1,wang2025tina} to finetune multilingual pretrained models like Llama \citep{grattafiori2024llama} or Qwen models \citep{yang2024qwen25}. There is a limited understanding how pretrained models' multilingual capability enables crosslingual reasoning of long CoTs, which our work aims to address.

\paragraph{Test-time scaling and s1} Test-time scaling is a new scaling paradigm where more computation budget is allocated for LLMs at inference time before committing to an answer \citep{snell2024scaling}. One example of test-time scaling is scaling up model generation length, particularly for RLMs with long CoTs capability \citep{snell2024scaling,goyal2024think,jaech2024openaio1,guo2025deepseek}. In particular, the s1 work \cite{muennighoff2025s1} demonstrates the effectiveness of a simple test-time scaling recipe: reasoning finetuning on small amount of training data with long CoTs (specifically 1k samples distilled from larger RLMs such as DeepSeek-R1) and scaling up inference budget at test time. Through test-time scaling of a 32B-parameter model, the authors achieve the state-of-the-art mathematical reasoning performance, and their models even rival industry-grade RLMs such as o1-mini \citep{jaech2024openaio1}. Nonetheless, similar to aforementioned RLMs literature, exploration of test-time scaling paradigm mostly evaluate on English math benchmarks \citep{snell2024scaling,ghosh2025multilingualmindsurvey,levi2024simple,xiang2025towards,wu2024inference,muennighoff2025s1}. Here, our work focuses on understanding how effective test-time scaling of English-centric RLMs, specifically s1 models, in multilingual settings for various domains.

\paragraph{Multilingual reasoning} Multilingual reasoning encompasses the ability of language models to perform complex reasoning tasks across different languages. Early work has demonstrated that chain-of-thought prompting in English can significantly improve performance on multilingual mathematical reasoning tasks \citep{shi2023mgsm}, which suggests that LLMs might rely on dominant languages like English as a pivot language for complex reasoning. 
Follow-up work explores several strategies such as translating the multilingual queries to English \citep{qin2023clp,zhu2024questiontranslate,ko2025understand}, aligning of latent representation spaces \citep{yoon2024langbridge,huang2024mindmerger} and reasoning outputs \citep{she2024mapo,yang2025languageimbalance} across languages, or expanding language coverage of reasoning training data \citep{chen2023breaking}. 
Our work focuses on understanding how controlling the length of long CoTs and their reasoning language at test time affects multilingual reasoning. 
One similar work \cite{son2025linguistic} experimented with controlling generation lengths of finetuned DeepSeek-R1-1.5B \cite{guo2025deepseek} but reported \textit{negative} results: increasing thinking tokens leads to minimal performance gains for mathematical reasoning in non-English languages. We believe that their negative findings are due to constrained model parameters, as we show that larger models \textit{can benefit} from crosslingual test-time scaling.

\section{Experimental Setup}\label{sec:experimental-setup}

\paragraph{Models} We use s1 models \cite{muennighoff2025s1} as our English-centric RLMs. In particular, we work with the s1.1 variants, which are multilingual Qwen2.5-Instruct models finetuned on 1k English-only reasoning data generated by DeepSeek-R1. 
We choose s1 models for two reasons: (1) their English mathematical reasoning capability reaches state-of-the-art performance, and (2) the models and the training data are fully open-sourced. 
We experiment with s1 models at different scales---specifically with 1.5B, 3B, 7B, 14B, and 32B parameters.

\paragraph{Budget forcing} Budget forcing refers to techniques for controlling inference budget for long CoTs \citep{muennighoff2025s1}, which can be done in two ways: (1) \textbf{truncation}, which \textit{cuts off} long CoTs after they reach maximum thinking tokens, or (2) \textbf{extrapolation}, which \textit{adds} tokens such as ``Wait'' at the end of CoTs to force the model continue reasoning. In our extrapolation setup, we experiment with adding ``Wait'' only once since we do not observe significant performance gains from lengthening CoTs.

\paragraph{Evaluation data} For research questions \RQone, \RQtwo, and \RQthree, we use the \textbf{Multilingual Grade School Math (MGSM)} benchmark \cite{shi2023mgsm}. It contains 250 grade-school math problems from the GSM8K dataset that have been manually translated into ten languages—Bengali (\texttt{bn}), German (\texttt{de}), Spanish (\texttt{es}), French (\texttt{fr}), Japanese (\texttt{ja}), Russian (\texttt{ru}), Swahili (\texttt{sw}), Telugu (\texttt{te}), Thai (\texttt{th}), and Mandarin Chinese (\texttt{zh}). We check for 8-gram overlap between s1’s training samples and MGSM (including the English subset) following standard practice \citep{zhang2024traintest}, and we observe no train–test overlap.

For \RQfour, addressed in~\cref{sec:cross-domain-generalization}, we evaluate on three different cross-domain benchmarks:
\begin{enumerate}[leftmargin=*]
  \item \textbf{Global-MMLU}~\citep{singh2024globalmmluunderstandingaddressing}: 
  a multilingual translated version of the original MMLU dataset~\citep{hendryckstest2021} that spans different subject topics, including STEM, \texttt{business}, humanities, \texttt{medical}, \texttt{social sciences}, etc. Additionally, it also contains annotations that identify whether the question is culturally agnostic or specific, allowing for more fine-grained analysis. 
  \item \textbf{Food ORiented cultural commonsense Knowledge (FORK)}~\citep{palta-rudinger-2023-fork}: a manually-curated set of commonsenseQA-style \citep{talmor-etal-2019-commonsenseqa} questions in English for probing cultural biases and assumptions on food-related customs around the world.
  \item \textbf{COPAL-ID}~\citep{wibowo-etal-2024-copal}: a manually-curated datasets following COPA's causal reasoning format \citep{roemmele2011choice} in Indonesian languages to evaluate causal reasoning with Indonesian cultural nuances.
\end{enumerate}

\begin{figure}[t]
    \centering
    \includegraphics[width=0.9\linewidth]{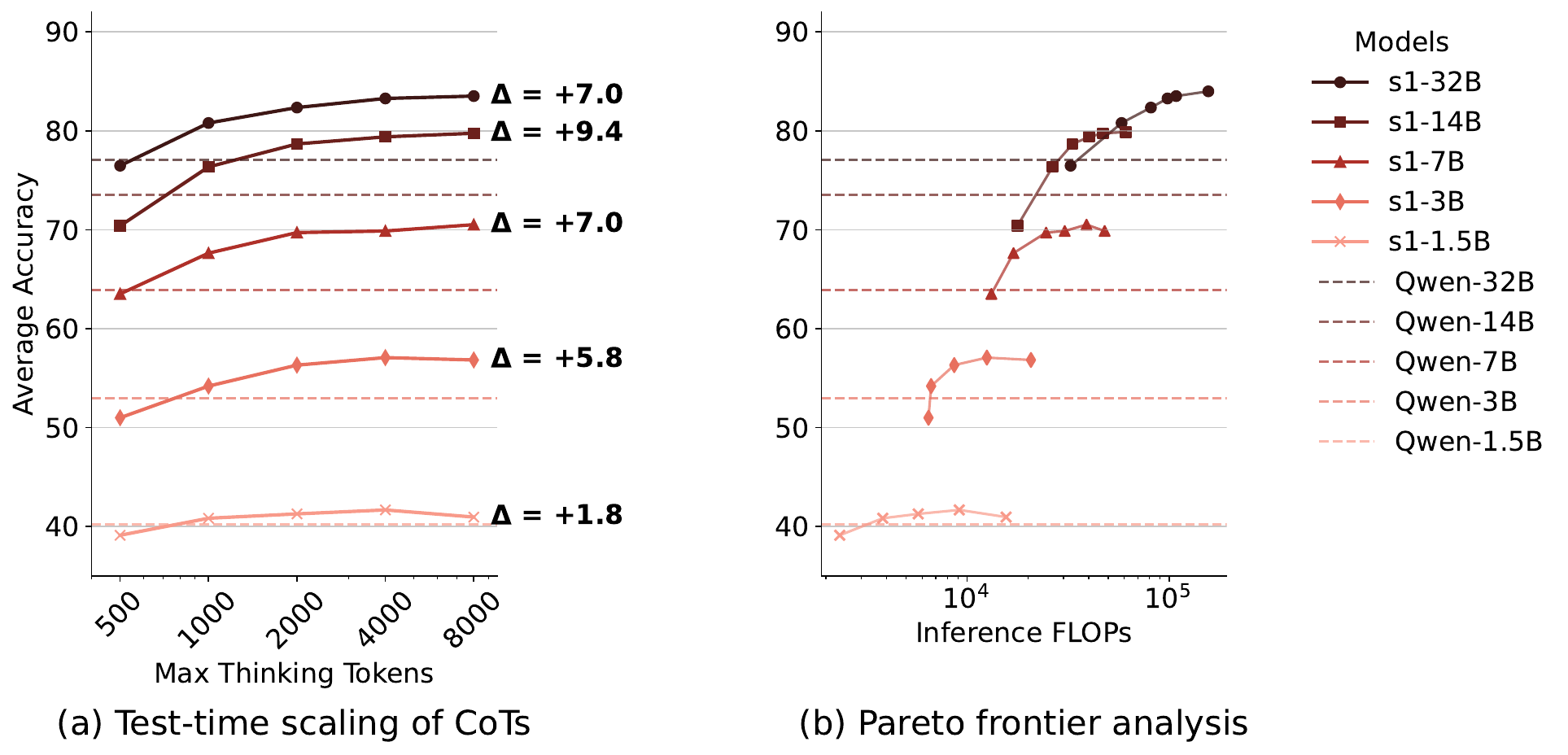}
    \caption{Crosslingual test-time scaling of s1 and Qwen models on the MGSM benchmark (\textit{excluding English}) across different model sizes. In subfigure (a) we enforce a hard limit of maximum thinking token, and in (b) we measure their inference FLOP compute for a Pareto frontier analysis. $\Delta$ measures the absolute difference between average accuracy scores at 0.5k and 8k maximum thinking tokens. Dash lines indicate the best few-shot prompting baseline performance of Qwen.}
    \label{fig:mgsm-trend}
\end{figure}

We use the \texttt{lm-evaluation-harness} library~\citep{eval-harness} as the main evaluation framework.\footnote{For cross-domain benchmarks, we use the configuration \texttt{generate\_until} tasks instead of the default \texttt{multiple\_choices} setting for cross-domain benchmarks and extract answers from generations using GPT-4o-mini~(see \Cref{app:mmlu-eval-details}). For Global-MMLU, we use the \texttt{default} subset of the benchmark and pick 7 languages that overlap with the MGSM benchmark.}. We evaluate with greedy decoding and report task accuracy, which is equivalent to pass@1.

\paragraph{Baselines} Qwen2.5-Instruct models, which are the base models of s1, are our main baseline models. For MGSM, we apply two prompting strategies. The first is zero-shot prompting, and the second is few-shot CoT prompting using the provided eight CoT examples in MGSM train set. For the latter, we follow \cite{shi2023mgsm} and experiment with prompting with few-shot CoT in English (EN-CoT) and in the same language as input prompt (native-CoT). In addition, for MGSM, we benchmark against prior state-of-the-art models  \citep{yue2023mammoth,she2024mapo,fan2025slam,yoon2024langbridge,team2025gemma3} and comparable English-centric RLMs such as R1-distilled-Qwen models \citep{guo2025deepseek} open-sourced by DeepSeek. For cross-domain benchmarks, we simply compare against zero-shot prompting of Qwen2.5-Instruct.

\section{Crosslingual Test-Time Scaling} \label{sec:crosslingual-mgsm}

In \RQone, we explore test-time scaling in a \textit{zero-shot crosslingual setting}, 
where English-centric reasoning models are applied to math problems in different languages. 
These models were finetuned from multilingual pretrained models in the same domain, therefore carrying high potential for crosslingual transfer.

\subsection{Effectiveness of Crosslingual Test-Time Scaling}
\paragraph{Crosslingual generalization of reasoning training and test-time scaling}
We report two main obsevations from \Cref{fig:mgsm-trend}~(a). First, we observe that s1 outperforms Qwen's few-shot prompting baseline across languages in MGSM (excluding English) when given high inference thinking budget. Second, crosslingual test-time scaling is effective for models with 3B parameters and above, with s1 at 14B size having the largest accuracy gain of $+\Delta9.4\%$ when maximum inference budget increases from 0.5k to 8k thinking tokens. We want to highlight that \textit{sufficient model capacity} is necessary for effective crosslingual test-time scaling, as test-time scaling only yields minimal benefits at 1.5B size. The model only experiences $+\Delta1.8\%$ accuracy gain with performance peak at 4000 maximum thinking tokens.

Our findings contrast the conclusion drawn by recent work \cite{son2025linguistic} that ``test-time scaling may not generalize as effectively to multilingual tasks.'' Our results suggest that the limitation observed by \cite{son2025linguistic} is due to their usage of 1.5B models in their experiments, instead of an inherent weakness of test-time scaling. Furthermore, \cite{son2025linguistic} posit that multilingual generalization of test-time scaling would occur for ``significantly larger'' models with at least 70B parameters, but we discover a substantially smaller parameter threshold at 3B parameters, above which models consistently benefit from crosslingual test-time scaling.

\paragraph{Pareto frontier analysis} \Cref{fig:mgsm-trend}~(b) illustrates the performance-efficiency trade-off across different sizes of the s1 model family on MGSM. We follow prior test-time scaling work \cite{snell2024scaling,sardana2023beyond} and compute the inference cost using the approximation $\text{FLOPs} = 2ND_\text{inference}$ where $N$ represents model
parameters and $D_\text{inference}$ the total number of
tokens generated at inference time, and we average across different languages. The figure further demonstrates the model capacity constraint on test-time scaling: The 32B and 14B models can reach accuracy levels that are simply unattainable for the smaller models, no matter how much test-time compute is allocated to those smaller models. 
Even at their maximum performance ceiling, the 3B and 7B models cannot match what the larger models achieve with moderate computational resources. 
While we observe accuracy-to-computation tradeoffs (i.e., better performance comes with using larger models and higher test-time compute), the 14B model offers a compelling compromise by achieving above 80\% accuracy with substantially lower inference FLOPs than the 32B model, representing a "sweet spot" on the Pareto frontier for practical applications.

\begin{table}[t]
\small
\caption{MGSM performance comparison against 14B-sized s1 model with maximum 8k thinking tokens. 
We report the language-breakdown accuracy from cited papers if available; otherwise, we reproduce using their open-sourced models without any inference budget constraint. We report the average length of the generations (avg. len) and the relative accuracy difference (\textcolor{green}{green text}) between s1-14B under extrapolation budget forcing and its baseline Qwen2.5-14B-Instruct. 
We \textbf{bold} both s1 performance and baseline models that outperform s1. 
}
\label{tab:14b-mgsm-performance}

\resizebox{\textwidth}{!}{%
\centering
\begin{tabular}{lr|rrrrrrrrrrrr}
\toprule
\textbf{Models} & avg len & \textbf{bn} & \textbf{de} & \textbf{en} & \textbf{es} & \textbf{fr} & \textbf{ja} & \textbf{ru} & \textbf{sw} & \textbf{te} & \textbf{th} & \textbf{zh} & \textbf{AVG} \\
\midrule
Qwen2.5-14B-Instruct \citep{yang2024qwen25} & 413.1 & 74.0 & 77.6 & 82.0 & 77.6 & 67.6 & 70.4 & 76.4 & 40.4 & 50.8 & 78.8 & 84.0 & 70.9 \\
 + 8-Shot EN-CoT \cite{shi2023mgsm} & 316.5 & 77.2 & 75.2 & 87.6 & 86.0 & 68.4 & 76.8 & 76.4 & 45.6 & 52.0 & 79.2 & 84.4 & 73.5 \\
 + 8-Shot Native-CoT \cite{shi2023mgsm} & 365.2 & 79.2 & 77.2 & 88.0 & 87.2 & 68.4 & 76.0 & 75.6 & 46.8 & 53.2 & 80.4 & 83.6 & 74.1\\

\noalign{\vspace{1pt}}
\hdashline
\noalign{\vspace{3pt}}
s1-14B (truncation) & 1912.9 & 82.0 & 84.8 & \textbf{92.8} & \textbf{88.4} & \textbf{85.2} & \textbf{83.6} & 86.8 & 55.6 & \textbf{59.6} & \textbf{85.2} & 86.4 & 80.9 \\

\textbf{s1-14B (extrapolation)} & 2352.3 & \textbf{82.8} & \textbf{86.8} & 92.4 & 86.4 & 83.2 & 83.2 & \textbf{88.8} & \textbf{57.2} & 58.0 & 84.8 & \textbf{87.6} & \textbf{81.0} \\

\ \ {Relative accuracy difference (\%)} & & \greenscriptsize{+11.9\%} & \greenscriptsize{+11.9\%} & \greenscriptsize{+12.7\%} & \greenscriptsize{+11.3\%} & \greenscriptsize{+23.1\%} & \greenscriptsize{+18.2\%} & \greenscriptsize{+16.2\%} & \greenscriptsize{+41.6\%} & \greenscriptsize{+14.2\%} & \greenscriptsize{+7.6\%} & \greenscriptsize{+4.3\%} & \greenscriptsize{+14.2\%} \\

\midrule
MetaMath-13B \citep{yu2024metamath} & 529.8 & 6.8 & 64.4 & 70.4 & 63.6 & 65.2 & 47.6 & 60.0 & 11.6 & 0.8 & 4.8 & 50.8  & 40.5 \\
MetaMathOctopus-13B \cite{she2024mapo} & 545.8 & 41.6 & 60.1 & 66.8 & 61.1 & 60.8 & 57.3 & 59.1 & 50.9 & 3.6 & 52.1 & 53.1 & 51.5 \\
MAPO-DPO-13B \cite{she2024mapo} & 552.4 & 54.7 & 69.5 & 70.5 & 70.6 & 71.3 & 69.0 & 68.2 & 62.9 & 4.0 & 64.7 & 68.2 & 61.2 \\
SL\textsc{am}-13B \cite{fan2025slam} & 101.5 & 45.6 & 62.8 & 71.2 & 67.6 & 65.2 & 54.0 & 64.4 & 46.4 & 2.4 & 47.6 & 58.8 & 53.3 \\
MetaMath-LB-15B \cite{yoon2024langbridge} & 93.2 & 50.0 & 63.6 & 67.6 & 63.2 & 61.6 & 42.0 & 60.0 & 41.6 & 36.4 & 52.8 & 48.0 & 53.5\\
MetaMath-LB-20B \cite{yoon2024langbridge} & 93.1 & 52.8 & 64.0 & 66.4 & 60.4 & 64.0 & 45.2 & 58.8 & 49.2 & 47.2 & 53.6 & 52.4 & 55.8 \\

\noalign{\vspace{1pt}}
\hdashline
\noalign{\vspace{3pt}}

R1-Distill-Qwen-14B \cite{guo2025deepseek} & 1030.7 & 66.0 & 77.2 & 83.6 & 80.4 & 74.4 & 78.4 & 82.4 & 22.4 & 22.4 & 74.8 & 79.6 & 67.4 \\
R1-Distill-Qwen-32B \cite{guo2025deepseek} & 1353.8 &  77.6 & 82.8 & 85.2 & 85.6 & 79.6 & 83.2 & 84.8 & 38.0 & 14.4 & 82.4 & 85.6 & 72.7 \\

Gemma-3-12B-it \cite{team2025gemma3} & 238.2 & 55.6 & 74.4 & 83.2 & 81.2 & 64.8 & 74.0 & 74.8 & \textbf{71.2} & \textbf{73.2} & 78.4 & 79.2 & 73.6 \\

Gemma-3-27B-it \cite{team2025gemma3} & 461.7 & 64.8 & 83.2 & 88.4 & 84.0 & 72.4 & 79.2 & 83.2 & \textbf{78.0} & \textbf{76.0} & 84.4 & 84.4 & 79.8 \\

Qwen3-14B \cite{team2025gemma3} & 1575.2 & \textbf{85.2} & 83.6 & \textbf{94.8} & \textbf{88.4} & \textbf{87.2} & 77.6 & \textbf{94.0} & \textbf{63.6} &  \textbf{80.0} & \textbf{86.8} & 85.2 & \textbf{84.2} \\

\bottomrule
\end{tabular}

    }

\end{table}

\subsection{Performance Comparison on MGSM Benchmark}

\paragraph{Comparison against Qwen baselines.} \Cref{tab:14b-mgsm-performance} shows that with cross-lingual test-time scaling, s1 gains substantial accuracy increase as compared to different baselines with Qwen2.5 models.
Furthermore, crosslingual test-time scaling benefits \textit{both} high-resource and low-resource languages. For instance, \texttt{fr} receives a significant $+\Delta23.1$\% relative accuracy increase, whereas \texttt{sw}---the worst-performing language for the base model Qwen---receives $+\Delta41.6\%$ relative accuracy improvement. Lastly, we observe similar performance for both truncation and extrapolation budget forcing strategies. This is because s1 models are already generating extensive reasoning chains so further lengthening the CoTs have minimal benefits. Similar trends of language-specific improvements are also observed in other s1 model sizes (\Cref{app:mgsm}).

\paragraph{Comparison against state-of-the-art models.} \Cref{tab:14b-mgsm-performance} shows that crosslingual test-time scaling of s1 models can serve as a strong multilingual baseline for MGSM, as it outperforms all prior state-of-the-art models that involve finetuning on multilingual data such as MetaMath, MAPO, and \cite{yu2024metamath,she2024mapo,fan2025slam,yoon2024langbridge}. We believe this is because these prior studies use Llama as their base models, which generate significantly shorter reasoning traces and lack sophisticated reasoning behaviors such as verification and backtracking compared to Qwen models \citep{gandhi2025cognitive}. 

Surprisingly, 14B-sized s1 can even outperform recent state-of-the-art reasoning models twice its size, namely DeepSeek's R1-Distill-Qwen-32B \cite{guo2025deepseek} and Google's Gemma-3-27B-it \cite{team2025gemma3}. We observe that R1-Distill-Qwen has substantially poorer performance on \texttt{sw} and \texttt{te}, suggesting that their 800k samples of English and Chinese training data \cite{guo2025deepseek} leads to \textit{catastrophic forgetting} of lower-resource languages. In contrast, s1 is only trained with 1k English samples for only 5 epochs \cite{muennighoff2025s1}, which leads to minimal forgetting and better crosslingual generalization. While the multilingual Gemma-3 models outperform s1 on low-resource languages, probably due to these languages being incorporated during reasoning finetuning, its performance gap against s1 on high-resource languages may be attributed to the shorter reasoning thinking time. 
Qwen3 \citep{qwen3_2025} is the most performant model, probably due to its long reasoning capability and extensive multilingual training data. 

\begin{CJK}{UTF8}{min}
\begin{example}[Example of \textit{quote-and-think} language-mixing pattern demonstrated by s1 where non-English phrases are mentioned in quotation marks and used for reasoning.]
{box:quote-and-think}

\textbf{Question}: ローブを作成するには、青色の繊維を2巻分、白色の繊維をその半分用いる必要があります。全体で何巻必要ですか？

(To make a robe, you need 2 rolls of blue fiber and half of the white fiber. How many rolls are needed total?)
\bigskip

\textbf{s1's Thinking}: Okay, let's see. ... The problem says "白色の繊維をその半分用いる" which is white fibers the half amount. "その" refers to the blue fibers, right? So half of the blue fibers, which is 2 rolls, so half is 1. So 2 + 1 is 3. Yeah, that seems right. Wait, sometimes these problems can be tricks, but I don't think so. If it had said "白色の繊維を2巻分とその半分" that would be different, but no. It's "青色の繊維を2巻分、白色の繊維をその半分". So white is half of blue. So calculation is straightforward. Blue 2, white 1, total 3. ...

\end{example}
\end{CJK}

\section{Language-Mixing Behaviors}
\label{sec:language-mixing-behaviors}

We notice that s1 models can mix languages in their CoT reasoning under inference scaling. Given a lack of systematic study of language mixing behaviors in crosslingual reasoning in prior work \citep{guo2025deepseek,ghosh2025multilingualmindsurvey}, we present an analysis of linguistic behaviors of s1 in multilingual math reasoning tasks to address \RQtwo. 

In particular, we are interested in two sub-questions of \RQtwo. First, how does the \textit{dominant language}––the primary language of generated response, also known as \textit{matrix language}––change after English-centric reasoning finetuning? Second, what are the language-mixing patterns exhibited by s1 in its reasoning outputs? We focus on four languages, namely \texttt{ja}, \texttt{ru}, \texttt{th}, and \texttt{zh}, and we refer our readers to \Cref{app:langmix-method} for detailed methodology.

\subsection{Dominant Language in Model Outputs}
\Cref{fig:mgsm_csw_dominant_lang} shows that s1 and its base model Qwen use entirely different dominant languages in their overall responses in multilingual settings. Qwen models
generate responses in the same language as the questions, especially for zero-shot and native CoT prompting settings. 
Surprisingly, even in crosslingual input settings where the 8-shot CoT reasoning samples are provided in English, Qwen models \textit{still stick to} generating outputs in the same language as the question. 
In contrast, s1 generates reasoning and final answers with English as the dominant language, which suggests that finetuning on only 1k English reasoning data can sufficiently change model behavior from multilingual to English-dominant. Similar behaviors are observed for models at smaller parameter sizes (\Cref{fig:mgsm_csw_dominant_lang_14B}).

\subsection{Language-Mixing Patterns During Reasoning}
\begin{figure}[t]
\begin{minipage}[t]{0.45\linewidth}
    \centering
    \includegraphics[width=\linewidth]{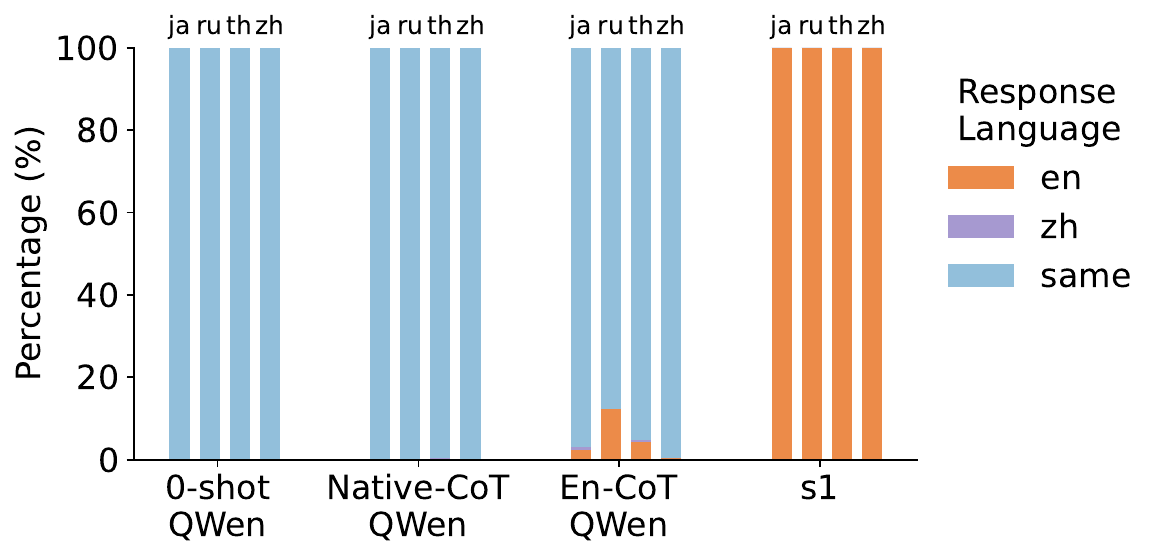}
    \caption{Proportion of dominant languages in models' entire responses when queried with multilingual math questions. ``same'' indicates that the response language is the same as query language.}
    \label{fig:mgsm_csw_dominant_lang}
\end{minipage} 
\hfill
\begin{minipage}[t]{0.45\linewidth}
    \includegraphics[width=\linewidth]{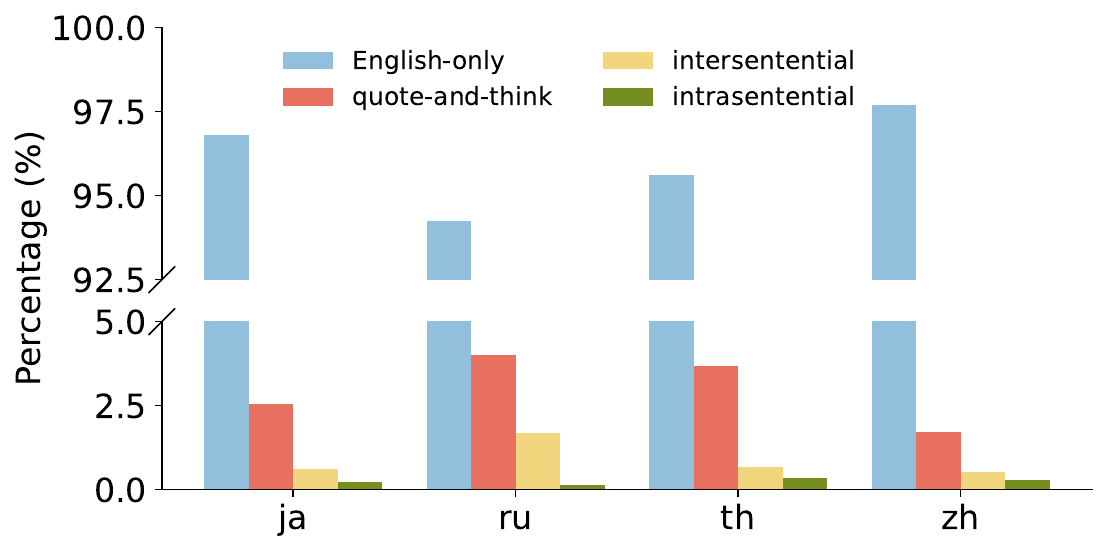}
    \caption{Breakdown of language-mixing patterns in s1's reasoning. Percentage indicates the probability of a sentence being English only, quoting non-English phrases (quote-and-think), entirely being in a different language (intersentential), or mixing different languages within the same sentence (intrasentential).}
    \label{fig:mgsm_csw_breakdown}
\end{minipage}
\end{figure}

\begin{CJK}{UTF8}{min}
\paragraph{Dominant mixing pattern} 
\Cref{fig:mgsm_csw_breakdown} shows that s1's reasoning is predominantly in English for all four languages of study. Specifically, at least 92.5\% of the sentences in s1's CoTs are in English only and do not mix languages. 
However, in the remaining cases when s1 mixes languages during reasoning, it primarily follows a sophisticated pattern to which we refer as \textbf{quote-and-think}. Particularly, s1 will first quote certain words or phrases, often from the input question, and then interpret their meanings and implications during its thinking process. This is demonstrated by the quoted phrase ``白色の繊維をその半分用いる'' and s1's literal translation ``white fibers the half amount'' in Box~\ref{box:quote-and-think}. In linguistics, this type of language-mixing is known as foreign-language quotation \citep{de2004foreign}. This language-mixing behavior happens due to crosslingual generalization of the quoting-and-thinking reasoning characteristic in s1's English finetuning data (see \Cref{app:s1.1-quoting}).

We want to emphasize that the \textit{quote-and-think pattern goes beyond simple translation.} As demonstrated in Box~\ref{box:quote-and-think}, s1 builds upon the extracted phrase and synthesizes a new multilingual setting where if the question had asked ``白色の繊維を2巻分とその半分'' (two and a half rolls of white fiber) it would have arrived at a different answer. Here, the model shows an understanding of how the syntactic structure in Japanese affects the semantic meaning of the math problem, which suggests that s1 is genuinely parsing and reasoning about the mathematical relationships expressed in Japanese and not merely translating the content to English before processing. This suggests that the multilingual capability of the base models is preserved for natural language understanding and allows s1 to reason about what it has understood about the question.

\end{CJK}

\paragraph{Other language-mixing patterns} \Cref{fig:mgsm_csw_breakdown} shows that, compared to other languages, Russian exhibits \textit{intersentential language-mixing}–––mixing of sentences of different languages–––the most, where English sentences are switched into Russian sentences and then back to English in the middle of the reasoning. This is undesirable \citep{guo2025deepseek} because it can create confusion for users expecting a coherent response in a single language and thus impacting readability. 
For \textit{intrasentential language-mixing}–––mixing of words of different languages in a single sentence–––we perform a deeper analysis (see \Cref{app:langmix-intra-breakdown}) and report that at least 70\% of cases resemble quote-and-think as the non-English phrases are extracted from the question prompt, but no quotation marks were provided. This further confirms that quote-and-think is the dominant pattern of language mixing in s1's reasoning process, even when not explicitly marked, and represents a natural strategy the model has developed to handle multilingual mathematical reasoning.

\section{Language Forcing}
\label{sec:language-forcing}
\begin{table}[t!]
\caption{Performance comparison of different language forcing strategies where we force s1 to reason in the same language as the question with 8k maximum thinking tokens. Languages are categorized into high-resource (HRL: \texttt{de}, \texttt{en}, \texttt{es}, \texttt{fr}, \texttt{ru}, \texttt{ja}, \texttt{zh}) and low-resource (LRL: \texttt{bn}, \texttt{sw}, \texttt{te}, \texttt{th}) groups. We use $\textcolor{green}{\uparrow}$ or $\textcolor{red}{\downarrow}$ to indicate if the language forcing method outperforms or underperforms the baseline of thinking in English.
}
\label{tab:lang_forcing}
    
    \centering
    \small
\scalebox{0.90}{
\begin{tabular}{ll|ccc|ccc}
    \toprule
     \multirow{2}{*}{\textbf{Model}}&\multirow{2}{*}{\textbf{Method}}&\multicolumn{3}{c|}{\textbf{Accuracy}} & \multicolumn{3}{c}{\textbf{Language Compliance}}\\
     &  & \textbf{ALL} & \textbf{HRL} & \textbf{LRL} & \textbf{ALL} & \textbf{HRL} & \textbf{LRL} \\
    \midrule
    \multirow{5}{*}{s1-32B} & \texttt{Baseline} & 87.4 & 91.2 & 80.8 & 0.0 & 0.0 & 0.0\\
    & \texttt{translated\_wait} & \textbf{87.8} $\textcolor{green}{\uparrow}$ & \textbf{91.3} $\textcolor{green}{\uparrow}$ & \textbf{81.7} $\textcolor{green}{\uparrow}$ & 19.4 & 30.4 & 0.2\\
    & \texttt{prefix} & 85.6 $\textcolor{red}{\downarrow}$ & \textbf{91.3} $\textcolor{green}{\uparrow}$ & 75.7 $\textcolor{red}{\downarrow}$ & 96.2 & 99.4 & 90.7\\
    & \texttt{system} & 86.0 $\textcolor{red}{\downarrow}$ & 90.3 $\textcolor{red}{\downarrow}$ & 79.3 $\textcolor{red}{\downarrow}$ & 47.7 & 71.4 & 6.2 \\
    & \texttt{combined} & 84.3 $\textcolor{red}{\downarrow}$ & 90.5 $\textcolor{red}{\downarrow}$ & 73.4 $\textcolor{red}{\downarrow}$ & \textbf{98.6} & \textbf{99.7} & \textbf{96.6}\\
    
    \midrule
    \multirow{5}{*}{s1-14B} & \texttt{Baseline} & 84.4 & 90.3 & 73.9 & 0.1 & 0.2 & 0.0\\
    & \texttt{translated\_wait} & \textbf{85.0} $\textcolor{green}{\uparrow}$ & 90.6 $\textcolor{green}{\uparrow}$ & \textbf{75.2} $\textcolor{green}{\uparrow}$ & 25.4 & 26.5 & 0.3\\
    & \texttt{prefix} & 82.0 $\textcolor{red}{\downarrow}$ & \textbf{90.7} $\textcolor{green}{\uparrow}$ & 66.9 $\textcolor{red}{\downarrow}$ & 96.4 & 99.6 & 89.2\\
    & \texttt{system} & 83.8 $\textcolor{red}{\downarrow}$ & 89.5 $\textcolor{red}{\downarrow}$ & 73.8 $\textcolor{red}{\downarrow}$ & 35.4 & 53.0 & 4.6\\
    & \texttt{combined} & 80.4 $\textcolor{red}{\downarrow}$ & 89.0 $\textcolor{red}{\downarrow}$ & 65.4 $\textcolor{red}{\downarrow}$ & \textbf{97.5} & \textbf{99.9} & \textbf{93.1}\\
    \midrule
    
    \multirow{5}{*}{s1-7B} & \texttt{Baseline} & \textbf{74.8} & \textbf{86.9} & \textbf{53.6} & 0.0 & 0.0 & 0.0\\
    & \texttt{translated\_wait} & 74.1 $\textcolor{red}{\downarrow}$ & 85.8 $\textcolor{red}{\downarrow}$ & \textbf{53.6} & 27.1 & 41.0 & 2.7\\
    & \texttt{prefix} & 69.5 $\textcolor{red}{\downarrow}$ & 84.0 $\textcolor{red}{\downarrow}$ & 44.2 $\textcolor{red}{\downarrow}$ & 89.8 & 96.9 & 77.3 \\
    & \texttt{system} & 73.1 $\textcolor{red}{\downarrow}$ & 84.8 $\textcolor{red}{\downarrow}$ & 52.7 $\textcolor{red}{\downarrow}$ & 70.5 & 96.6 & 24.9\\
    & \texttt{combined} & 69.9 $\textcolor{red}{\downarrow}$ & 84.5 $\textcolor{red}{\downarrow}$ & 44.4 $\textcolor{red}{\downarrow}$ & \textbf{96.0} & \textbf{98.0} & \textbf{92.6}\\
    \bottomrule
\end{tabular}
}
\end{table}

When a multilingual user interacts with LLMs, it is natural to expect the LLMs to respond in the language consistent with the user's query. Therefore, in \RQthree, we are interested in understanding if we can perform \textit{language forcing}–––controlling an English-centric RLM to generate reasoning in a particular language---and if the difference in reasoning language affects performance.

\subsection{Methodology: Language Forcing Techniques}
We experiment with the following language forcing techniques to control s1's reasoning language:
\begin{itemize}[leftmargin=*]
\item \textbf{Translated Wait} (\texttt{translated\_wait}): Building upon extrapolation budget forcing strategy that explicitly extends reasoning traces \citep{muennighoff2025s1}, once the model finishes English reasoning, we append a translated ``Wait'' token as an intervention strategy to force the model to switch language and continue reasoning in our chosen language.

\item \textbf{Prefix} (\texttt{prefix}): We appended a prefix string translation-equivalent of ``Okay, let me try to figure this out.'' at the beginning of the reasoning generation in order to guide the model's generation in our chosen language. We also apply the \texttt{translated\_wait} strategy and append the translated ``Wait'' token.

\item \textbf{System Prompt} (\texttt{system}): We use a system prompt to control the language use in model generation. Specifically, we translate the system prompt ``You are a helpful assistant.''\footnote{We remove the part of ``You are Qwen, created by Alibaba Cloud.'' because English proper nouns like `Qwen' and `Alibaba' do not have translation equivalents in many non-English languages.} into our chosen language and add the translation-equivalent of the instruction ``You must think and answer only in \{language\}''.

\item \textbf{Combined} (\texttt{combined}): This method uses all the techniques above to maximize control over the model's reasoning language.
\end{itemize}

We experiment with language forcing on the MGSM benchmark with $K=11$ languages. Given a question in $k \in K$ language (i.e., query language), we force the model to reason in $k' \in K$ language (i.e., reasoning language). We compute two evaluation metrics: task accuracy and \textit{language compliance}. The former follows the setup described in~\Cref{sec:experimental-setup}, whereas the latter measures the proportion of tokens generated in the intended reasoning language relative to tokens produced in other languages. In other words, language compliance quantifies the effectiveness of s1 in following explicit language constraints. To only measure the effect of reasoning language and eliminate the confounding factors such as introduction of \texttt{prefix} string, our baseline is to let s1 models think in its dominant language English using the \texttt{combined} setup.

\subsection{In-Language Forcing}\label{sec:in-language-forcing}
In-language forcing refers to the setting where the reasoning language matches the query language, i.e., $k' = k$. Here, we focus on analyzing if we should perform in-language forcing for English-centric RLMs or let models reason in their natural reasoning language (i.e., English for s1). We showcase an example of s1 thinking in Japanese due to language forcing in \Cref{app:lf-example}.

\paragraph{Reasoning in high-resource languages}
As shown in \Cref{tab:lang_forcing}, reasoning in high-resource languages (HRLs) achieve similar scores–––typically within a 1-2 points difference compared to baseline–––regardless of language forcing strategy. One possible explanation is that these languages are highly represented within pretraining of Qwen, which allows for effective crosslingual transfer from English reasoning finetuning. 

\paragraph{Reasoning in low-resource languages}
\Cref{tab:lang_forcing} highlights that, in contrast to HRLs, for low-resource languages (LRLs), \texttt{combined} strategy substantially underperforms the baseline. This suggests that English reasoning finetuning does not transfer to LRLs as effectively, and thus s1 is relatively incapable of reasoning in LRLs. Strategies that permit a mix of English and the target language---particularly \texttt{translated\_wait} and \texttt{system}---outperform strict in-language approaches such as \texttt{combined}. This is because s1 can lean on its English reasoning ability while incorporating the target language, which appears to be beneficial in scenarios with limited language-specific training data. For example, in (te) and (sw), the 14B and 32B models using \texttt{translated\_wait} consistently outperform those using \texttt{prefix} or \texttt{combined} strategies by 4–5 accuracy points (see \Cref{app:detailed_results_in_lang} for detailed results). 

\paragraph{Trade-off between accuracy and language compliance} 
\Cref{tab:lang_forcing} shows that we can successfully control the reasoning language of RLMs in their long CoTs, but that usually come at a cost of task performance. 
For instance, for both 32B- and 14B-sized s1 models, the \texttt{combined} strategy achieves near 100\% language compliance for both HRLs and LRLs, but it yields the worst accuracy compared to other language forcing strategies.
For 7B-sized models, we observe that we need a combination of all strategies (i.e., \texttt{combined}) to achieve high language compliance. This suggests that it is non-trivial to control the reasoning language, as different choices of language forcing strategy can significantly affect how consistently s1 adhere to our specified reasoning language.

\subsection{Crosslingual Language Forcing}
\definecolor{color1}{HTML}{7682D3}   %
\definecolor{color2}{HTML}{8BA0EB}
\definecolor{color3}{HTML}{A3BCFB}
\definecolor{color4}{HTML}{BBD2FF}
\definecolor{color5}{HTML}{D3E1F8}
\definecolor{color6}{HTML}{E7E6E6}   %
\definecolor{color7}{HTML}{F6DBCD}
\definecolor{color8}{HTML}{F9C5B0}
\definecolor{color9}{HTML}{F3A995}
\definecolor{color10}{HTML}{E2867C}
\definecolor{color11}{HTML}{CA4F67}  %

\begin{table}[t]
\caption{Performance scores across different reasoning languages given query language. We use 11 color codes to \textit{rank each row} to highlight the high- (\textcolor{color1}{blue}) and low-performing (\textcolor{color11}{red}) reasoning language given a query language. We also \textbf{bold} the best-performing reasoning language. Lastly, we use $^\searrow$ to indicate the average accuracy when the reasoning language is the same as query language (i.e., average of the diagonals).}
\label{tab:NxN}
    \centering
    \scalebox{0.85}{
        \begin{tabular}{l|ccccccccccc|c}
        \toprule
        \textbf{Query}&\multicolumn{11}{c|}{\textbf{Reasoning Language}} & \textbf{Range}\\
         \textbf{Language}& \textbf{bn} & \textbf{de} & \textbf{en} & \textbf{es} & \textbf{fr} & \textbf{ja} & \textbf{ru} & \textbf{sw} & \textbf{te} & \textbf{th} & \textbf{zh} & \textbf{(max - min)} \\
        \midrule
        bn & \cellcolor{color7}79.2 & \cellcolor{color2}85.2 & \cellcolor{color1}\textbf{86.8} & \cellcolor{color3}84.4 & \cellcolor{color5}81.6 & \cellcolor{color6}81.2 & \cellcolor{color4}83.6 & \cellcolor{color11}62.4 & \cellcolor{color10}75.6 & \cellcolor{color8}80.8 & \cellcolor{color6}81.2 & 24.4\\
        de & \cellcolor{color7}88.4 & \cellcolor{color5}89.2 & \cellcolor{color2}90.4 & \cellcolor{color6}88.8 & \cellcolor{color1}\textbf{90.8} & \cellcolor{color3}90.0 & \cellcolor{color8}87.6 & \cellcolor{color11}75.6 & \cellcolor{color10}78.4 & \cellcolor{color7}88.0 & \cellcolor{color4}89.6 & 15.2\\
        en & \cellcolor{color8}93.2 & \cellcolor{color5}94.4 & \cellcolor{color5}94.4 & \cellcolor{color2}95.2 & \cellcolor{color3}94.8 & \cellcolor{color5}94.4 & \cellcolor{color8}93.2 & \cellcolor{color10}84.0 & \cellcolor{color10}84.0 & \cellcolor{color3}94.8 & \cellcolor{color1}\textbf{96.8} & 12.8\\
        es & \cellcolor{color9}86.4 & \cellcolor{color4}92.4 & \cellcolor{color1}\textbf{93.6} & \cellcolor{color1}\textbf{93.6} & \cellcolor{color4}92.4 & \cellcolor{color7}90.8 & \cellcolor{color3}93.2 & \cellcolor{color11}76.6 & \cellcolor{color10}82.8 & \cellcolor{color8}90.0 & \cellcolor{color7}90.8 & 17.0\\
        fr & \cellcolor{color7}87.2 & \cellcolor{color7}87.2 & \cellcolor{color2}88.4 & \cellcolor{color7}87.2 & \cellcolor{color4}88.0 & \cellcolor{color1}\textbf{89.6} & \cellcolor{color2}88.4 & \cellcolor{color11}72.8 & \cellcolor{color10}77.6 & \cellcolor{color7}87.2 & \cellcolor{color4}88.0 & 16.8\\
        ja & \cellcolor{color9}79.2 & \cellcolor{color2}84.8 & \cellcolor{color4}83.6 & \cellcolor{color7}81.6 & \cellcolor{color1}\textbf{85.6} & \cellcolor{color6}82.0 & \cellcolor{color2}84.8 & \cellcolor{color10}71.6 & \cellcolor{color11}74.0 & \cellcolor{color1}\textbf{85.6} & \cellcolor{color4}83.6 & 14.0\\
        ru & \cellcolor{color9}89.2 & \cellcolor{color5}91.2 & \cellcolor{color3}92.4 & \cellcolor{color8}89.6 & \cellcolor{color1}\textbf{93.6} & \cellcolor{color4}92.0 & \cellcolor{color3}92.4 & \cellcolor{color11}77.6 & \cellcolor{color10}80.8 & \cellcolor{color7}90.0 & \cellcolor{color5}91.2 & 16.0\\
        sw & \cellcolor{color8}45.6 & \cellcolor{color2}58.8 & \cellcolor{color1}\textbf{59.6} & \cellcolor{color4}55.2 & \cellcolor{color3}55.6 & \cellcolor{color7}47.6 & \cellcolor{color6}48.4 & \cellcolor{color9}44.4 & \cellcolor{color11}32.4 & \cellcolor{color8}45.2 & \cellcolor{color5}52.0 & 27.2\\
        te & \cellcolor{color7}53.2 & \cellcolor{color4}56.4 & \cellcolor{color1}\textbf{60.0} & \cellcolor{color4}56.4 & \cellcolor{color1}\textbf{60.0} & \cellcolor{color3}57.2 & \cellcolor{color5}55.2 & \cellcolor{color11}34.8 & \cellcolor{color6}54.4 & \cellcolor{color8}53.6 & \cellcolor{color9}52.8 & 25.2\\
        th & \cellcolor{color9}80.8 & \cellcolor{color4}88.4 & \cellcolor{color3}89.2 & \cellcolor{color4}88.4 & \cellcolor{color1}\textbf{91.2} & \cellcolor{color7}87.2 & \cellcolor{color7}87.2 & \cellcolor{color11}66.4 & \cellcolor{color10}69.2 & \cellcolor{color8}86.4 & \cellcolor{color2}88.8 & 24.8\\
        zh & \cellcolor{color9}85.2 & \cellcolor{color7}86.8 & \cellcolor{color3}89.6 & \cellcolor{color6}87.2 & \cellcolor{color7}86.8 & \cellcolor{color4}88.8 & \cellcolor{color1}\textbf{90.8} & \cellcolor{color11}73.6 & \cellcolor{color10}77.2 & \cellcolor{color8}86.0 & \cellcolor{color2}89.2 & 17.2\\
        \midrule
        AVG & \cellcolor{color9}78.9 & \cellcolor{color3}83.2 & \cellcolor{color1}\textbf{84.4} & \cellcolor{color5}82.6 & \cellcolor{color2}83.7 & \cellcolor{color7}81.9 & \cellcolor{color6}82.3 & \cellcolor{color11}67.3 & \cellcolor{color10}71.5 & \cellcolor{color8}80.7 & \cellcolor{color4}82.2 & {$^\searrow$}81.2 \\
        \bottomrule
    \end{tabular}
    }
\end{table}

Given our success in controlling the reasoning language of s1 with \texttt{combined} strategy, we explore if there is a particular language that is best served as reasoning language for s1. Particular, for each query language $m \in M$, we force the model to all $M$ possible languages, resulting in an exhaustive $M\times M$ query-reasoning language-pair analysis.\footnote{This analysis is
is computationally heavy, so we only focus on 14B-sized s1 models.}

\paragraph{Performance comparison of reasoning languages} \Cref{tab:NxN} shows that reasoning in HRLs such as \texttt{en}, \texttt{fr}, or \texttt{de} yield similarly high performance (the accuracy difference is within 1 to 2 points), with English being the most performant reasoning language and French being the close second. We discover two surprising findings: first, even though the Qwen2.5 base model is highly pretrained in Chinese \citep{yang2024qwen25}, it is not necessarily the best reasoning language, even when the question is asked in \texttt{zh}; second, neither reasoning in \texttt{en} nor in query language necessarily yields the best performance---quite the contrary, even reasoning in languages that are usually less represented in pretraining data \cite{joshi-etal-2020-state} such as \texttt{ru} and \texttt{th} can achieve the best performance for query languages in other families such as \texttt{ja}.
Lastly, we observe that languages that are considered as slightly less-resourced \citep{joshi-etal-2020-state} such as \texttt{th} and \texttt{bn} still achieve nearly 80\% overall accuracy, but further lower-resourced languages such as \texttt{sw} or \texttt{te} result in substantially lower overall accuracy. 

\begin{figure}[t]
    \centering
    \includegraphics[width=\linewidth]{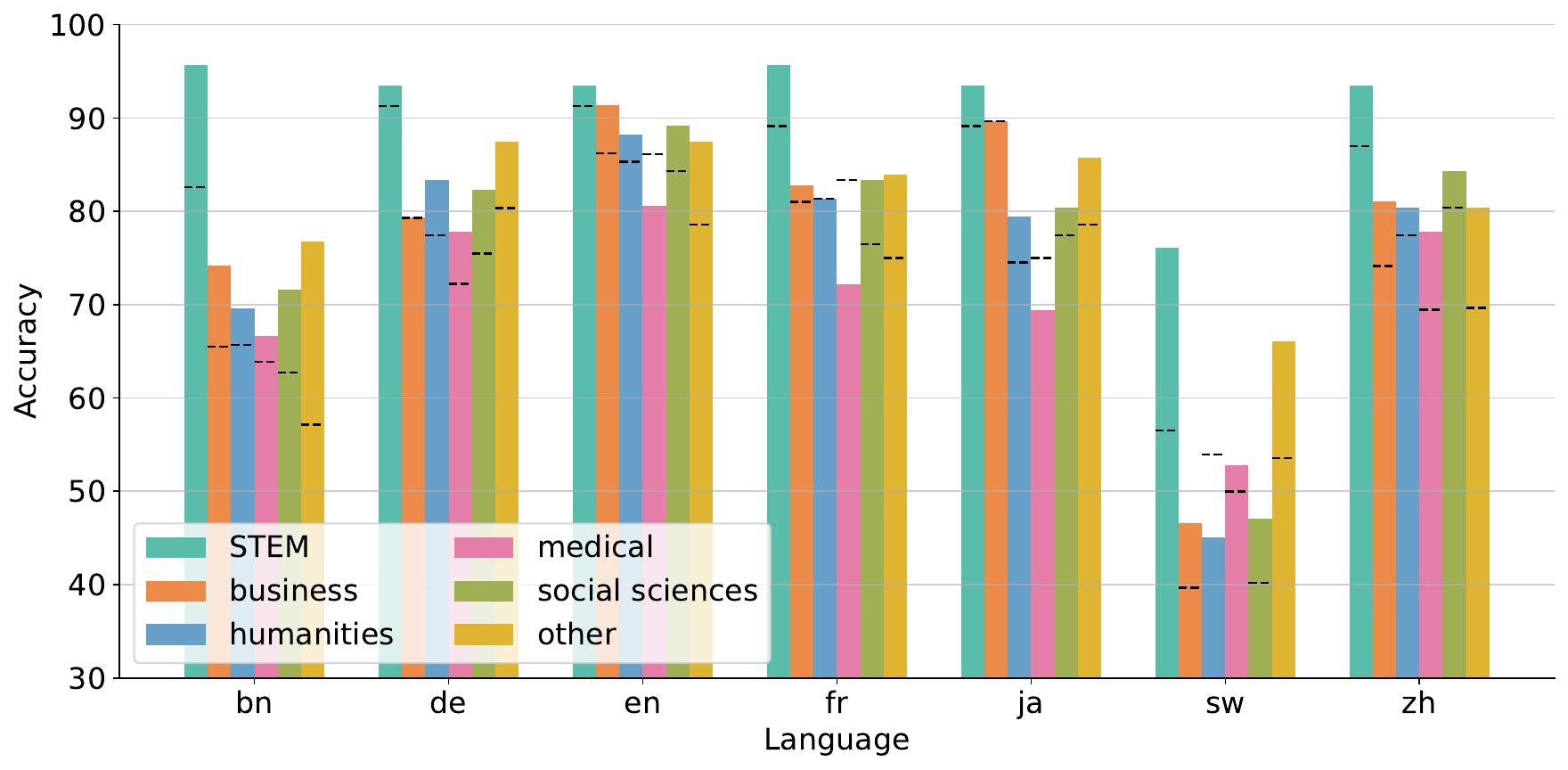}
    \caption{Language and domain breakdown for Global-MMLU benchmark. Dashed lines indicate the performance of zero-shot prompting of Qwen-32B-Instruct models.}
    \label{fig:mmlu_lang}
\end{figure}

\begin{wrapfigure}{r}{0.5\textwidth}
  \begin{center}
    \includegraphics[width=0.48\textwidth]{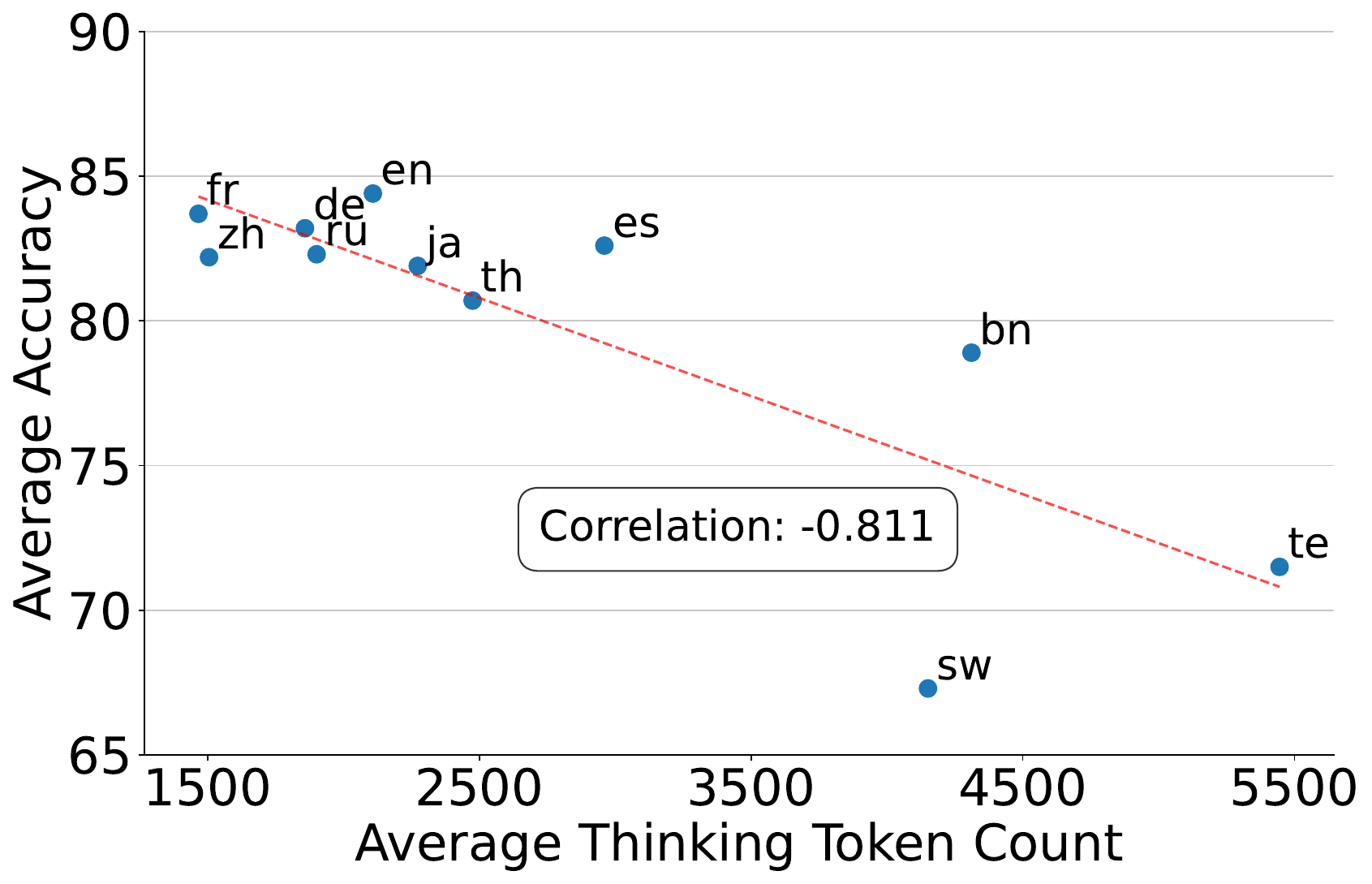}
  \end{center}
  \caption{MGSM accuracy against number of thinking tokens in s1 models' outputs in different reasoning languages.}
  \label{fig:acc_vs_token}
\end{wrapfigure}

\paragraph{Inference cost analysis} Our analysis of inference costs across reasoning languages in \Cref{fig:acc_vs_token} reveals a significant negative correlation (-0.811) between token count and mathematical problem-solving accuracy. Reasoning in LRLs not only underperform their HRL counterpart (with accuracy below 80\%), but they also demand substantially more computational resources at test-time---often exceeding 3,000 tokens in their long CoTs for \texttt{bn}, \texttt{sw}, and \texttt{te}. This token efficiency disparity at test-time translates directly to computational costs; for instance, reasoning in Swahili requires approximately 3.5 times more compute than French for the same tasks. 
This is very likely due to the well-known \textit{tokenization disparity} for LRLs, where tokenized lengths for an equivalent sentence in different languages are much larger for LRLs \citep{petrov2023language,ahia2023do}.

\paragraph{Choice of query language} \Cref{tab:NxN} sheds light on whether we should translate inputs into HRLs such as English for reasoning tasks, which has proven to be an effective strategy \citep{qin2023clp,zhu2024questiontranslate}. Our results are consistent with prior work: merely translating the question from Swahili to French can boost the accuracy from 59.6 to 90.8 even when the model reasons in French---a language that s1 is not trained to reason in. Besides, based on the range column, which measures difference between the best and worst reasoning languages for a particular query language, the model is less sensitive to query language in HRLs than in LRLs as exhibited by the smaller range. In other words, querying s1 in HRLs increases the model's consistency in achieving the same correct answer with different reasoning languages.

\section{Cross-Domain Generalization}\label{sec:cross-domain-generalization}

Since s1 models obtain strong crosslingual math performance with English-only training, a natural question to ask is whether such generalization extends to other non-math domains that may require knowledge recall or cultural reasoning. We address this research question \RQfour using the cross-domain benchmarks described in \Cref{sec:experimental-setup}.

\subsection{In-domain generalization}
 
\begin{figure}
    \centering
    \includegraphics[width=0.9\linewidth]{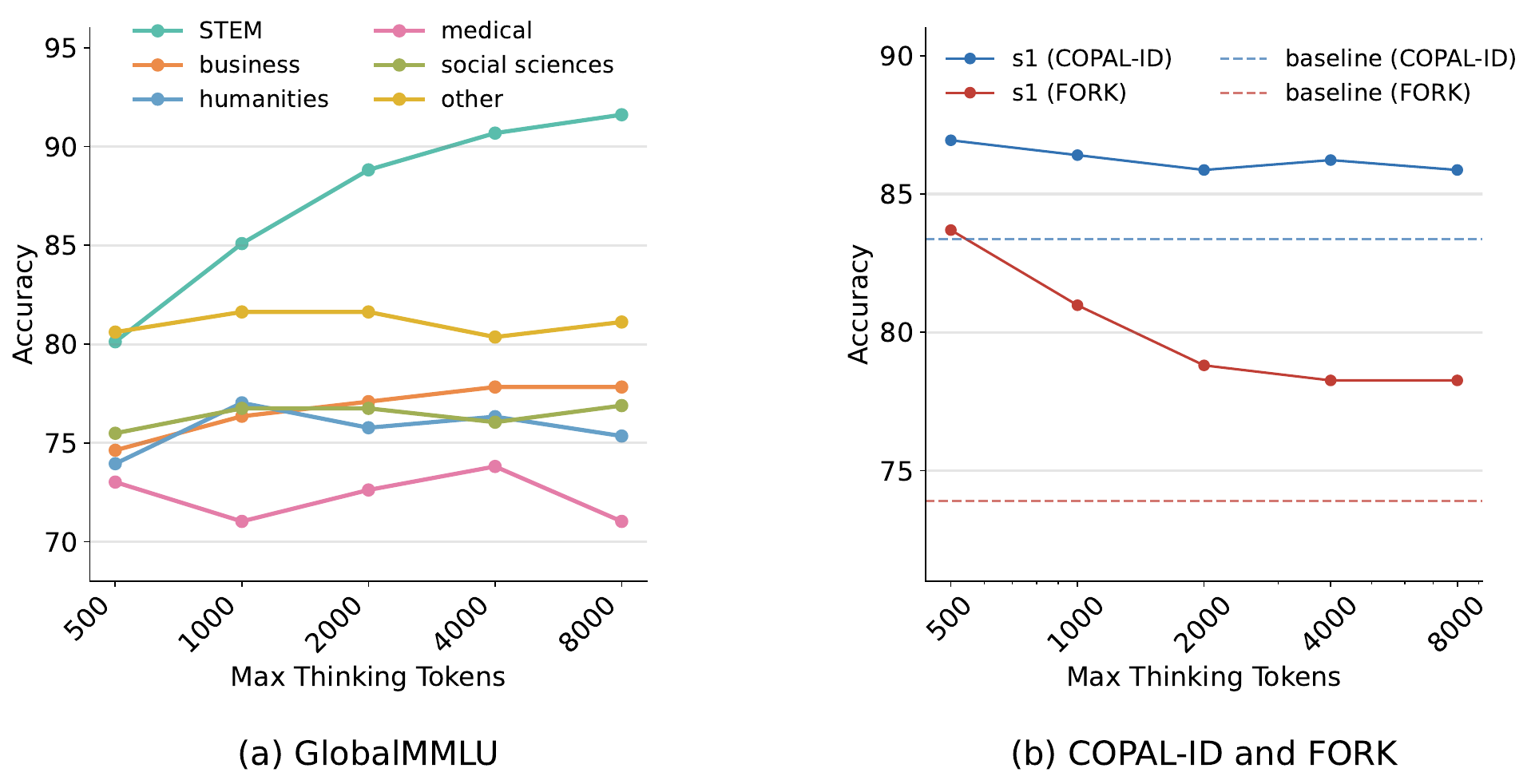}
    \caption{Effects of thinking time for s1 models on different domains of Global-MMLU benchmark (subfigure (a)) and cultural commonsense knowledge (FORK) and reasoning (COPAL-ID) benchmarks (subfigure (b)). Similar to \Cref{fig:mmlu_lang}, we added dashed lines as zero-shot prompting of Qwen-32B as baselines in (b).}
    \label{fig:cross_domain_scaling}
\end{figure}

\paragraph{STEM domains} \Cref{fig:cross_domain_scaling}~(a) shows that performance on STEM subject domain in Global-MMLU improves with test-time scaling of thinking tokens (cyan line). Furthermore, s1 outperforms its base model Qwen2.5-32B-Instruct (black dashed line for cyan blocks in \Cref{fig:mmlu_lang}) by a large margin, especially for \texttt{bn} and \texttt{sw} where both gain $+\Delta19.8\% $ $(56.5\%\rightarrow76.1\%)$ and $+\Delta13.1\%$ $(82.6\%\rightarrow95.7\%)$ accuracy respectively. 
Given that s1 training data includes OlympicArena dataset \citep{huang2024olympicarena} that encompasses various STEM subject knowledge such as biology and astronomy, our results here further corroborates \Cref{sec:crosslingual-mgsm} findings on \textbf{strong crosslingual in-domain generalization}.

\subsection{Out-of-domain generalization}

\paragraph{non-STEM domains} We report two main observations. First, \Cref{fig:mmlu_lang} shows that the benefits of reasoning finetuning are domain-specific. For instance, \texttt{business} and \texttt{social sciences} domains gain slight improvements, but the \texttt{medical} domain experiences up to nearly 10\% decrease in accuracy even for HRLs such as \texttt{en} and \texttt{fr} compared to the baseline Qwen model. Second, we report \textbf{minimal cross-domain generalization of test-time scaling} from \Cref{fig:cross_domain_scaling}~(a). Domains such as medicine do not benefit from scaling up thinking tokens, as increasing maximum thinking tokens from 0.5k to 4k tokens merely improves accuracy by only $+\Delta0.8\% $ $(73.0\%\rightarrow73.8\%)$, and further scaling to 8000 thinking tokens even reduces accuracy by $-\Delta2.0\% $ $(73.0\%\rightarrow71.0\%)$. Out of all non-STEM domains, \texttt{business} benefits the most from test-time scaling ($+\Delta3.2\%$), but the accuracy gain still lags behind STEM domain ($+\Delta11.5\%$) by a huge margin.

\paragraph{Cultural-specific knowledge and reasoning}

For cultural commonsense knowledge and reasoning benchmarks, we observe similar findings that there is minimal benefits of test-time scaling of s1. \Cref{fig:cross_domain_scaling}~(b) shows that while reasoning finetuning improves overall model performance over Qwen baselines (dashed lines), scaling up test-time thinking compute does not improve performance. In fact, for the English FORK benchmark, increasing thinking tokens leads to substantially poorer performance. This is also known as \textit{overthinking} \citep{liu2024mind} where reasoning models expend excessive compute in their long CoTs and lead to worse performance \citep{cuadron2025danger,chen2024not,sui2025stop}.

\begin{figure}[t]
    \centering
    \includegraphics[width=0.7\linewidth]{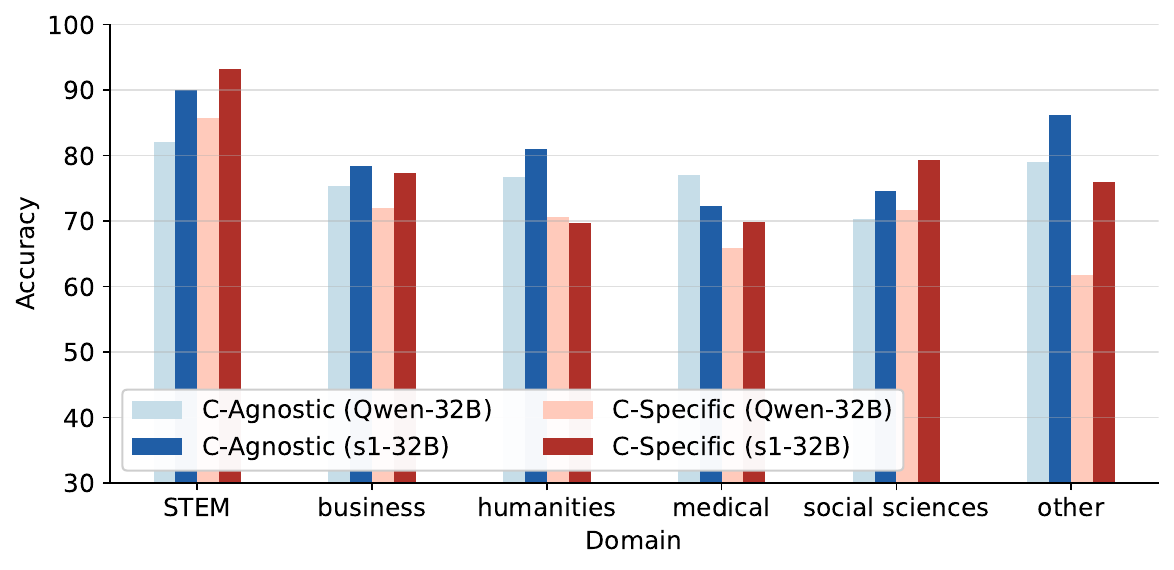}
    \caption{Performance comparison between s1 and zero-shot prompting of Qwen baseline for culturally-agnostic (C-Agnostic) and culturally-specific (C-Specific) questions in different domains. Results are average across languages.}
    \label{fig:mmlu_ca_cs}
\end{figure}

As for Global MMLU, which contains additional annotations for whether questions belong to culture-specific or culture-agnostic types, Figure~\ref{fig:mmlu_ca_cs} shows that there is no obvious pattern explaining which type benefits the most from the s1-training. For the STEM category, the improvements are almost equivalent. However, in other domains, we observe inconsistent performance change. For instance, 
we observe that performance drops in the \texttt{humanities} subject mostly belong to the culture-specific type while the opposite for the \texttt{\texttt{medical}} domain. 
We also experiment with language forcing (using the \texttt{combined} strategy from \Cref{sec:language-forcing}) to see if reasoning in the same language as the question helps with performance, but we observe minimal and often inconsistent performance gains from language forcing (see \Cref{app:mmlu-language-forcing} for more details).

\section{Discussion and Future Work}
\paragraph{Data-efficient English reasoning finetuning} One notable finding from our work is that s1 remains capable of generating text in different languages and experiences minimal catastrophic forgetting---a phenomenon where the model loses its ability to generate fluent text in other languages after language-specific supervised finetuning \citep{yong-etal-2023-bloom,kotha2024understanding}. 
In contrast, R1-Distill-Qwen baseline experiences significant catastrophic forgetting for low-resource languages. 
This suggests that data-efficient finetuning with a small number of reasoning finetuning steps (s1 is only trained with 1k English samples for 5 epochs) is advisable for English-centric reasoning finetuning to preserve multilingual capability.

\paragraph{Reasoning finetuning with multilingual data} 
Given the limitations of crosslingual test-time scaling–––poor reasoning in low-resource languages and in cross-domain settings–––one potential solution would be to curate multilingual reasoning training data with wide language and domain coverage. However, the current practice of generating multilingual reasoning training data simply through translation technology \citep{chen2023breaking,son2025linguistic} is insufficient, as it is well-established that translation models including LLMs still suffer from poor cultural alignment and Western-centric bias \citep{singh-etal-2024-translating,rystrom2025multilingual} as well as poor translation performance with low-resource languages \citep{robinson-etal-2023-chatgpt,court-elsner-2024-shortcomings,song2025llm}. Future work should systematically explore the effectiveness of multilingual augmentation techniques such as back-translation or synthetic data generation \citep{whitehouse-etal-2023-llm,yong-etal-2024-lexc}.

\paragraph{Reasoning in low-resource languages (LRLs)} 
Our results indicate poor performance of using LRLs as reasoning language.
Furthermore, unfairness in tokenization for LRLs \citep{petrov2023language} leads to significantly higher inference costs for test-time scaling of RLMs. This disparity creates accessibility barriers for multilingual communities, as deploying reasoning models in LRLs becomes prohibitively expensive. Future work should focus on developing more equitable tokenization strategies for reasoning across diverse languages \citep{liang-etal-2023-xlmv,han2025adapters,xue-etal-2022-byt5}.

\paragraph{Crosslingual generalization for small RLMs} Our observation of successful crosslingual test-time scaling happening at 3B parameter threshold corroborates contemporary work that shows small models below 3B parameters struggle to learn from long CoTs \cite{li2025small} and generalize in multilingual settings \citep{son2025linguistic}. While there are ongoing work creating English-centric RLMs smaller than 3B size \citep{wang2025tina}, we argue that future work on small RLMs should benchmark on multilingual reasoning tasks rather than solely on English data. This would not only advance the development of multilingual reasoning models but also help identify factors that enable crosslingual reasoning.

\section{Conclusion}
Our work studies the effectiveness of test-time scaling of English-centric reasoning language models for multilingual math reasoning, and we perform pareto frontier analyses of different model sizes and inference thinking compute. Furthermore, we document the language-mixing patterns naturally exhibited by reasoning models and report the novel ``quote-and-think'' pattern. We also investigate the effects of the choice of reasoning languages as well as query languages, demonstrating that successful generalization of English reasoning finetuning to other high-resource languages. Finally, we observe minimal cross-domain generalization of test-time scaling, thus highlighting the need for future work to explore more robust and domain-agnostic reasoning strategies, particularly for multilingual applications.

\begin{ack}
We thank Nihal Nayak and Peilin Yu from BATS research group at Brown University for helpful discussions. Disclosure: Stephen Bach is an advisor to Snorkel AI, a company that provides software and services for data-centric artificial intelligence.
\end{ack}

\renewcommand{\bibsection}{\section*{References}}
\setcitestyle{numbers,square}

\bibliography{main}

\appendix

\newpage
\tableofcontents

\newpage
\section*{Appendix}

\section{Authors' Contributions}\label{app:author-contributions}

Zheng-Xin Yong led the project and ran most of the experiments. Muhammad Farid Adilazuarda and Jonibek Mansurov worked on language forcing. Ruochen Zhang worked on cross-domain generalization with GlobalMMLU. Niklas Muennighoff trained s1 models at different model sizes. 

Alham Fikri Aji, Stephen Bach, Julia Kreutzer, Genta Indra Winata, and Niklas Muennighoff advised the project at the early stage. All authors contributed to the paper writing.

\section{Further Details on Crosslingual Test-Time Scaling} \label{app:mgsm}

\Cref{tab:mgsm-s1-32b}, \Cref{tab:mgsm-s1-7b}, \Cref{tab:mgsm-s1-3b}, and \Cref{tab:mgsm-s1-1.5b} shows the s1 performance against Qwen baselines on the MGSM benchmark. Relative accuracy difference measures the relative gains for s1 under extrapolation budget forcing compared to zero-shot prompting Qwen models (first row), except for \Cref{tab:mgsm-s1-1.5b} where the relative accuracy is measured for s1 under truncation budget forcing.

\begin{table}[htbp]
\centering
\caption{MGSM performance comparison against 32B-sized s1 model with maximum 8k thinking tokens.}
\label{tab:mgsm-s1-32b}
\resizebox{\textwidth}{!}{%
\centering
\begin{tabular}{lr|rrrrrrrrrrrr}
\toprule
\textbf{Models} & avg len & \textbf{bn} & \textbf{de} & \textbf{en} & \textbf{es} & \textbf{fr} & \textbf{ja} & \textbf{ru} & \textbf{sw} & \textbf{te} & \textbf{th} & \textbf{zh} & \textbf{AVG} \\
\midrule
Qwen-32B-Instruct & 365.6 & 82.0 & 79.6 & 84.0 & 80.0 & 69.2 & 80.8 & 77.2 & 54.0 & 56.4 & 84.4 & 84.0 & 75.6 \\
 + 8-Shot EN-CoT & 264.7 & 82.0 & 80.4 & 89.6 & 84.8 & 66.8 & \textbf{85.2} & 77.6 & 56.8 & 55.6 & 84.8 & 84.8 & 77.1 \\
 + 8-Shot Native-CoT & 179.9 & 82.4 & 78.0 & 90.0 & 87.2 & 66.8 & 80.8 & 74.0 & 57.2 & 59.6 & 87.2 & 88.0 & 77.4 \\

\noalign{\vspace{1pt}}
\hdashline
\noalign{\vspace{3pt}}

s1-32B (truncation) & 1682.1 & \textbf{89.2} & 86.8 & \textbf{94.8} & \textbf{90.0} & \textbf{84.8} & 84.8 & \textbf{91.2} & 66.8 & 65.6 & 87.2 & \textbf{88.8} & 84.5 \\

\textbf{s1-32B (extrapolation)} & 2610.1 & \textbf{89.2} & \textbf{88.8} & 93.2 & 89.6 & 83.6 & 84.8 & 88.8 & \textbf{70.8} & \textbf{68.4} & \textbf{88.0} & 88.0 & \textbf{84.8} \\

\ \ {Relative accuracy difference (\%)} & & \greenscriptsize{+8.8\%} & \greenscriptsize{+11.6\%} & \greenscriptsize{+11.0\%} & \greenscriptsize{+12.0\%} & \greenscriptsize{+20.8\%} & \greenscriptsize{+5.0\%} & \greenscriptsize{+15.0\%} & \greenscriptsize{+31.1\%} & \greenscriptsize{+21.3\%} & \greenscriptsize{+4.3\%} & \greenscriptsize{+4.8\%} & \greenscriptsize{+12.2\%} \\

\bottomrule
\end{tabular}
}
\end{table}

\begin{table}[htbp]
\centering
\caption{MGSM performance comparison against 7B-sized s1 model with maximum 8k thinking tokens.}
\label{tab:mgsm-s1-7b}
\resizebox{\textwidth}{!}{%
\centering
\begin{tabular}{lr|rrrrrrrrrrrr}
\toprule
\textbf{Models} & avg len & \textbf{bn} & \textbf{de} & \textbf{en} & \textbf{es} & \textbf{fr} & \textbf{ja} & \textbf{ru} & \textbf{sw} & \textbf{te} & \textbf{th} & \textbf{zh} & \textbf{AVG} \\
\midrule
Qwen-7B-Instruct & 537.7 & 59.2 & 69.2 & 78.0 & 72.8 & 66.4 & 67.2 & 71.2 & 13.6 & 33.2 & 68.8 & 79.6 & 61.7 \\
 + 8-Shot EN-CoT & 537.8 & 62.0 & 72.4 & 86.4 & 78.8 & 62.4 & 68.0 & 76.8 & 15.2 & 33.2 & 70.4 & 76.8 & 63.9 \\
 + 8-Shot Native-CoT & 480.1 & 65.2 & 74.4 & \textbf{90.4} & 76.4 & 65.2 & 71.6 & 68.8 & 18.4 & 20.8 & 69.6 & 76.8 & 63.4 \\

\noalign{\vspace{1pt}}
\hdashline
\noalign{\vspace{3pt}}

\textbf{s1-7B (truncation)} & 3767.1 & 65.2 & 82.8 & 88.8 & \textbf{86.0} & 82.0 & \textbf{78.8} & \textbf{86.4} & \textbf{21.6} & \textbf{38.8} & \textbf{80.0} & \textbf{83.6} & \textbf{72.2} \\

s1-7B (extrapolation) & 4363.5 & \textbf{70.8} & \textbf{84.0} & \textbf{90.4} & 83.6 & \textbf{84.4} & 74.8 & 84.4 & 19.2 & 36.4 & 78.4 & 82.8 & 71.7 \\

\ \ {Relative accuracy difference (\%)} & & \greenscriptsize{+19.6\%} & \greenscriptsize{+21.4\%} & \greenscriptsize{+15.9\%} & \greenscriptsize{+14.8\%} & \greenscriptsize{+27.1\%} & \greenscriptsize{+11.3\%} & \greenscriptsize{+18.5\%} & \greenscriptsize{+41.2\%} & \greenscriptsize{+9.6\%} & \greenscriptsize{+14.0\%} & \greenscriptsize{+4.0\%} & \greenscriptsize{+16.2\%} \\
\bottomrule
\end{tabular}
}
\end{table}

\begin{table}[htbp]
\centering
\caption{MGSM performance comparison against 3B-sized s1 model with maximum 8k thinking tokens.}
\label{tab:mgsm-s1-3b}
\resizebox{\textwidth}{!}{%
\centering
\begin{tabular}{lr|rrrrrrrrrrrr}
\toprule
\textbf{Models} & avg len & \textbf{bn} & \textbf{de} & \textbf{en} & \textbf{es} & \textbf{fr} & \textbf{ja} & \textbf{ru} & \textbf{sw} & \textbf{te} & \textbf{th} & \textbf{zh} & \textbf{AVG} \\
\midrule
Qwen-3B-Instruct & 1023.3 & 37.6 & 58.8 & 74.0 & 66.0 & 54.4 & 54.8 & 64.8 & 9.2 & 7.6 & 56.8 & 68.4 & 50.2 \\
 + 8-Shot EN-CoT & 281.3 & 48.0 & \textbf{67.6} & 79.2 & 71.2 & 65.2 & 58.0 & 70.4 & \textbf{12.4} & 14.8 & 60.4 & 68.4 & 56.0 \\
 + 8-Shot Native-CoT & 1657.0 & 36.0 & 63.2 & 80.0 & 70.8 & 58.4 & 52.0 & 62.0 & 9.6 & 9.6 & 59.2 & 70.4 & 51.9\\

\noalign{\vspace{1pt}}
\hdashline
\noalign{\vspace{3pt}}

s1-3B (truncation) & 4813.3 & \textbf{56.8} & 66.8 & \textbf{82.0} & 74.4 & 69.6 & 60.4 & 72.4 & 10.4 & 16.8 & \textbf{68.8} & 72.0 & 59.1 \\

\textbf{s1-3B (extrapolation)} & 5367.1 & 55.2 & 65.6 & 81.6 & \textbf{76.4} & \textbf{71.6} & \textbf{60.8} & \textbf{74.4} & 9.6 & \textbf{20.0} & \textbf{68.8} & \textbf{74.0} & \textbf{59.8} \\

\ \ {Relative accuracy difference (\%)} & & \greenscriptsize{+46.8\%} & \greenscriptsize{+11.6\%} & \greenscriptsize{+10.3\%} & \greenscriptsize{+15.8\%} & \greenscriptsize{+31.6\%} & \greenscriptsize{+10.9\%} & \greenscriptsize{+14.8\%} & \greenscriptsize{+4.3\%} & \greenscriptsize{+163.2\%} & \greenscriptsize{+21.1\%} & \greenscriptsize{+8.2\%} & \greenscriptsize{+19.1\%} \\

\bottomrule
\end{tabular}
}
\end{table}

\begin{table}[htbp]
\centering
\caption{MGSM performance comparison against 1.5B-sized s1 model with maximum 8k thinking tokens. We didn't run extrapolation budget forcing since without it, s1 already generates extremely long CoTs.}
\label{tab:mgsm-s1-1.5b}
\resizebox{\textwidth}{!}{%
\centering
\begin{tabular}{lr|rrrrrrrrrrrr}
\toprule
\textbf{Models} & avg len & \textbf{bn} & \textbf{de} & \textbf{en} & \textbf{es} & \textbf{fr} & \textbf{ja} & \textbf{ru} & \textbf{sw} & \textbf{te} & \textbf{th} & \textbf{zh} & \textbf{AVG} \\
\midrule
Qwen-1.5B-Instruct & 2991.7 & 10.4 & 35.6 & 66.0 & 52.8 & 41.2 & 31.2 & 43.6 & \textbf{2.0} & 1.2 & 31.2 & 56.0 & 33.7\\
 + 8-Shot EN-CoT & 1100.3 & 21.6 & 46.4 & \textbf{70.0} & 58.0 & 55.2 & 37.2 & 51.6 & \textbf{2.8} & \textbf{6.4} & 41.2 & 52.0 & 40.2 \\
 + 8-Shot Native-CoT & 1729.9 & 14.0 & 44.4 & \textbf{71.6} & 52.4 & 41.6 & 34.4 & 39.2 & \textbf{3.2} & 1.6 & 33.2 & 54.0 & 35.4 \\

\noalign{\vspace{1pt}}
\hdashline
\noalign{\vspace{3pt}}

s1-1.5B (truncation) & 8227.2 & \textbf{27.6} & \textbf{51.2} & 66.8 & \textbf{62.8} & \textbf{56.0} & \textbf{43.2} & \textbf{55.6} & 1.6 & \textbf{6.4} & \textbf{46.4} & \textbf{58.8} & \textbf{43.3} \\

\ \ {Relative accuracy difference (\%)} & & \greenscriptsize{+165.4\%} & \greenscriptsize{+43.8\%} & \greenscriptsize{+1.2\%} & \greenscriptsize{+18.9\%} & \greenscriptsize{+35.9\%} & \greenscriptsize{+38.5\%} & \greenscriptsize{+27.5\%} & \redscriptsize{-20.0\%} & \greenscriptsize{+433.3\%} & \greenscriptsize{+48.7\%} & \greenscriptsize{+5.0\%} & \greenscriptsize{+28.5\%} \\
\bottomrule
\end{tabular}
}
\end{table}

\section{Further Details on Language-Mixing Behaviors}

\subsection{Methodology}\label{app:langmix-method}

To filter out language-mixed sentences, we first identify the \textit{dominant language}, also known as matrix language, of the generated response using the state-of-the-art language identification library \texttt{lingua}. Then, we use the NLP library \texttt{stanza} to perform sentence segmentation according to the matrix language and obtain individual sentences. Finally, we use \texttt{lingua} to annotate the language label of each sentence and of each individual word token in the sentence. 

We classify language-mixing patterns into three categories: (1) quote-and-think, where words or phrases in foreign language are quoted in quotation marks; (2) intersentential, where the entire sentence is in a language entirely different from generation dominant language, and (3) intrasentential, where words, phrases or clauses of different languages are present in the same sentence. We refer our readers to \Cref{app:langmix-annos} for our annotation procedures.

We focus on four languages, namely Japanese (ja), Russian (ru), Thai (th), and Mandarin Chinese (zh), as they are readily supported by the libraries \texttt{stanza} and \texttt{lingua}. We avoid languages with Latin scripts due to their possible shared vocabulary with English and results in incorrect language classification. For instance, the German word `\textit{also}' (therefore) in the sentence ``Sie isst 3 Eier zum Frühstück und verwendet 4 Eier für Muffins, also verwendet sie insgesamt 3 + 4 = 7 Eier pro Tag.'' was misclassified as English word. 

\subsection{Annotation for Language-Mixing Patterns in s1 Reasoning}\label{app:langmix-annos}

We collect the language label with the highest probability assigned to \textit{the entire sentence}, and we label a sentence belonging to ``intersentential'' language-mixing for s1 if the sentence is non-English, as the dominant language of s1's overall output is English. We then check language labels for individual word tokens. If there are mixing of different languages \textit{within the same sentence}, and quotation marks are present around the non-English words or phrases, then the sentence is assigned with the ``quote-and-think'' label. Otherwise, if quotation marks are not present, the sentence is assigned with the ``intrasentential'' label.

\subsection{Dominant Language in 14B-Sized Model Outputs}
\begin{figure}[h]
    \centering
    \includegraphics[width=0.6\linewidth]{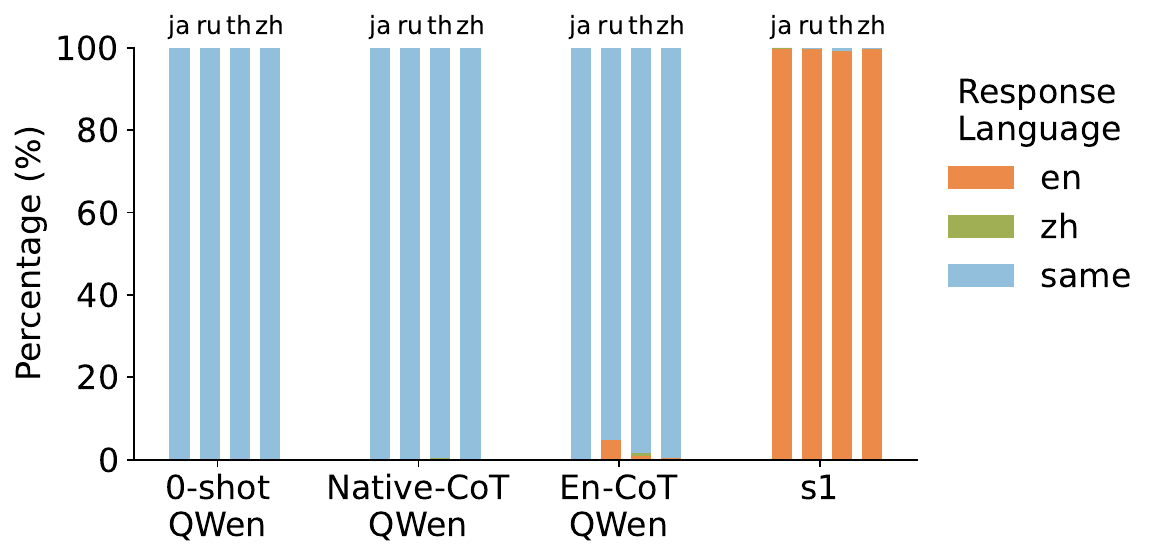}
    \caption{Proportion of dominant languages used by 14B-sized models' responses when queried with Japanese (ja), Russian (ru), Thai (th), and Mandarin Chinese (zh) languages. ``same'' indicates that the response language is the same as query language.}
    \label{fig:mgsm_csw_dominant_lang_14B}
\end{figure}

\Cref{fig:mgsm_csw_dominant_lang_14B} shows the dominant language distribution in model outputs when MGSM questions are asked in Japanese (ja), Russian (ru), Thai (th), and Mandarin Chinese (zh).

\subsection{Quote-And-Think Pattern in s1's Training Data}\label{app:s1.1-quoting}

Among 1k English training samples of s1 models, 68.3\% of the samples exhibit the quote-and-think pattern, among which at least half of them involves directly copying from the question prompts. This suggests that the quote-and-think language-mixing pattern is due to crosslingual transfer of the original s1 model's learned behavior of quoting phrases from question prompts during its long CoTs thinking process.

\subsection{Fine-Grained Analysis of s1-32B's Intrasentential Language Mixing}\label{app:langmix-intra-breakdown}

\begin{figure}[h]
    \centering
    \includegraphics[width=0.6\linewidth]{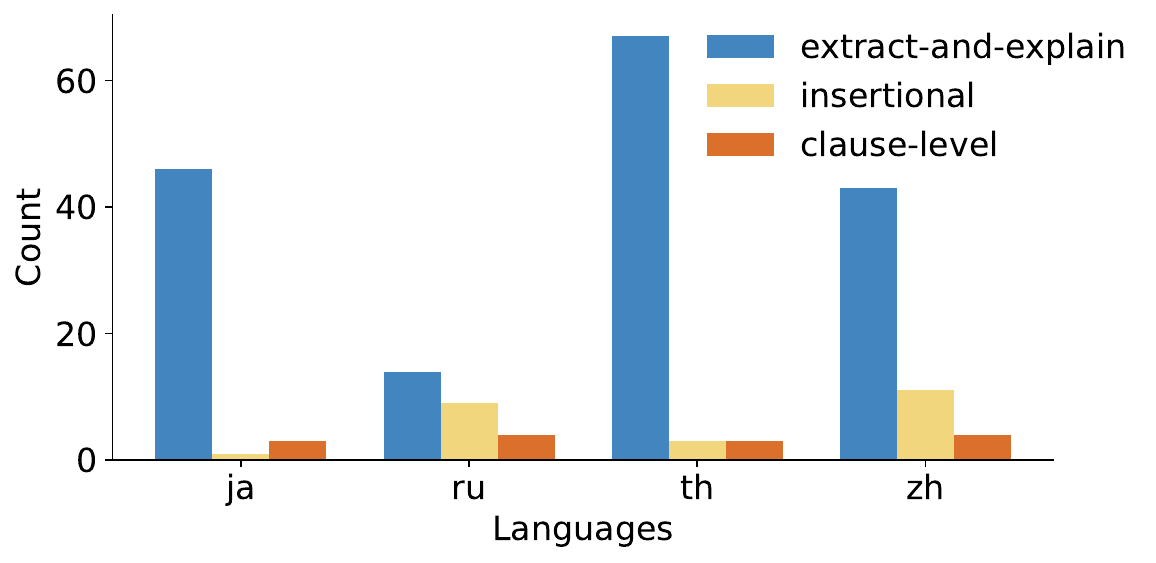}
    \caption{Breakdown of s1-32B's finegrained intrasentential language mixing patterns.}
    \label{fig:mgsm_csw_intra_32B}
\end{figure}

We perform human annotations on the intrasententially language-mixed sentences during reasoning and classify if each sentence belongs to one of the following categories: (1) \textbf{extract-and-explain}, where the non-English phrases are taken directly from the original input prompt but \textit{without} quotation marks given (this resembles quote-and-think but no quotation marks are generated around the non-English phrases); (2) \textbf{insertional code-switching}, where non-English lexical items (usually nouns) are inserted into the morphosyntactic frame of the English sentence (an example would be ``I want to eat \textit{nasi goreng}'' where \textit{nasi goreng} is a Malay word for fried rice), and (3) \textbf{clause-level code-switching}, where switching between two languages within a single sentence structure at clausal level (an example would be ``I want to go to the library \textit{dan bersedia untuk peperiksaan}'' where \textit{dan bersedia untuk peperiksaan} is a Malay clause for ``and prepare for the exam''.) \Cref{fig:mgsm_csw_intra_32B} demonstrates the distribution of each category, with extract-and-explain being the dominant language-mixing patterns.

\section{Further Details on Language Forcing}
\label{app:detailed_results}

\subsection{Example of CoTs after Language Forcing}\label{app:lf-example}

\begin{CJK}{UTF8}{min}
\begin{example}[Example of CoTs prdocued from \texttt{combined} language forcing strategy.]
{box:lf}

\textbf{Question}: ローブを作成するには、青色の繊維を2巻分、白色の繊維をその半分用いる必要があります。全体で何巻必要ですか？

(To make a robe, you need 2 rolls of blue fiber and half of the white fiber. How many rolls are needed total?)
\bigskip

\textbf{s1's Thinking}: よし、解いてみよう。問題はローブを作るために必要な繊維の巻数を計算するものだね。青色の繊維が2巻必要で、白色の繊維はその半分、つまり1巻必要だ。だから合計で3巻か。でも待って、問題文をもう一度確認しよう。
\\
\\
「ローブを作成するには、青色の繊維を2巻分、白色の繊維をその半分用いる必要があります。全体で何巻必要ですか？」そうだ、青色が2巻、白色がその半分。つまり青色の半分は1巻だから、白色は1巻必要。だから合計3巻。これでいいのかな。でも、もしかして青色の2巻と白色の半分を足すと、2 + 0.5 = 2.5巻って意味じゃないのだろうか。でも問題の言い回しが少しややこしいな。
\\
\\
待って、原文をもう一度。「青色の繊維を2巻分、白色の繊維をその半分用いる」というのは、青色が2巻、そして白色が「その半分」。つまり、「青色の半分」なのか、それとも「2巻の半分」なのか。日本語の文法で考えると、「その半分」は直前の数詞、「2巻分」を指している可能性が高い。つまり、青色2巻、白色は2巻の半分なので1巻。だから合計3巻。でも、もし「青色の繊維を2巻分、白色の繊維をその半分」というのが、青色が2巻、白色が青色の半分、つまり1巻、という意味なら、同じく3巻。結局同じ結果になるのか。
\\
\\
それに、問題が「全体で何巻必要ですか？」と聞いているので、必要な総巻数を聞かれているのだろう。つまり、青色2巻、白色1巻、合計3巻。これが答えだと思われる。でも、もしこの問題が英語だと、\"how many rolls are needed in total?\" 
\\
だと仮定すると、3 rolls が答え。 でも、もしかして青色2巻、白色半分、つまり2.5巻という解釈もある。でも、問題文の日本語の表現からすると、青色2巻、白色1巻、合計3巻が妥当だと思う。
\\
...

\end{example}
\end{CJK}

Box~\ref{box:lf} show the example of CoTs generated by s1 from the \texttt{combined} language-forcing strategy. This example contrasts Box~\ref{box:quote-and-think} where Box~\ref{box:quote-and-think} showcases s1's natural CoTs without language forcing. 

Note that we still observe the sophisticated \textit{quote-and-think} pattern where here, English phrases are quoted for reasoning (the dominant language is Japanese). Specifically, in that sentence where quote-and-think occurs, it is translated to `if this question was in English, it would be ``how many rolls are needed in total?'''

\subsection{Language Compliance}
\label{app:detailed_results_lang_consist}
Detailed results for language compliance can be found in Tables \ref{tab:lang-compliance-32B}, \ref{tab:lang-compliance-14B} and \ref{tab:lang-compliance-7B} for 32B, 14B and 7B, respectively. 
\begin{table}[t]
\caption{Results for \textbf{s1-32B}.}
\centering
\resizebox{\textwidth}{!}{%
\begin{tabular}{l|l|r|l|r}
\toprule
\textbf{Strategy} & \textbf{Reasoning Language} & \textbf{avg len} & \textbf{Language Distribution} & \textbf{Language Compliance} \\ 
\midrule
\multirow{11}{*}{\texttt{translated\_wait}}    & bn & 3073.3 & ENGLISH: 99.6, BENGALI: 0.4 & 0.4 \\
& de & 2405.3 & ENGLISH: 51.6, GERMAN: 48.4 & 48.4 \\
& en & 1833.1 & ENGLISH: 100.0 & 100.0 \\
& es & 2401.5 & ENGLISH: 88.4, SPANISH: 11.6 & 11.6 \\
& fr & 2379.7 & ENGLISH: 90.8, FRENCH: 8.8, CHINESE: 0.4 & 8.8 \\
& ja & 2515.1 & ENGLISH: 70.8, JAPANESE: 29.2 & 29.2 \\
& ru & 2601.1 & ENGLISH: 90.4, RUSSIAN: 9.2, CHINESE: 0.4 & 9.2 \\
& sw & 2611.4 & ENGLISH: 100.0 & 0.0 \\
& te & 3821.3 & ENGLISH: 99.6, TELUGU: 0.4 & 0.4 \\
& th & 1894.7 & ENGLISH: 99.6, CHINESE: 0.4 & 0.0 \\
& zh & 1776.3 & ENGLISH: 94.4, CHINESE: 5.6 & 5.6 \\
\midrule
\multirow{11}{*}{\texttt{prefix}}   & bn & 3320.0 & BENGALI: 98.8, ENGLISH: 1.2 & 98.8 \\
& de & 1747.2 & GERMAN: 98.8, CHINESE: 1.2 & 98.8 \\
& en & 1729.7 & ENGLISH: 100.0 & 100.0 \\
& es & 2790.6 & SPANISH: 99.6, ENGLISH: 0.4 & 99.6 \\
& fr & 1822.9 & FRENCH: 100.0 & 100.0 \\
& ja & 2321.6 & JAPANESE: 98.8, ENGLISH: 1.2 & 98.8 \\
& ru & 1564.2 & RUSSIAN: 98.8, ENGLISH: 0.8, CHINESE: 0.4 & 98.8 \\
& sw & 3083.9 & SWAHILI: 91.2, ENGLISH: 8.4, JAPANESE: 0.4 & 91.2 \\
& te & 6912.7 & TELUGU: 95.2, ENGLISH: 4.8 & 95.2 \\
& th & 2126.3 & THAI: 77.6, CHINESE: 20.4, ENGLISH: 2.0 & 77.6 \\
& zh & 1150.1 & CHINESE: 99.6, ENGLISH: 0.4 & 99.6 \\
\midrule
\multirow{11}{*}{\texttt{system}}   & bn & 2693.8 & ENGLISH: 90.4, BENGALI: 9.6 & 9.6 \\
& de & 1979.3 & GERMAN: 100.0 & 100.0 \\
& en & 1728.3 & ENGLISH: 100.0 & 100.0 \\
& es & 2241.6 & ENGLISH: 69.6, SPANISH: 30.4 & 30.4 \\
& fr & 2346.0 & ENGLISH: 68.4, FRENCH: 31.2, CHINESE: 0.4 & 31.2 \\
& ja & 1805.7 & JAPANESE: 99.2, ENGLISH: 0.8 & 99.2 \\
& ru & 2180.0 & ENGLISH: 59.6, RUSSIAN: 39.2, CHINESE: 1.2 & 39.2 \\
& sw & 2721.2 & ENGLISH: 98.8, SWAHILI: 0.8, CHINESE: 0.4 & 0.8 \\
& te & 3869.3 & ENGLISH: 93.6, TELUGU: 6.4 & 6.4 \\
& th & 1930.3 & ENGLISH: 90.8, THAI: 8.0, CHINESE: 1.2 & 8.0 \\
& zh & 1162.1 & CHINESE: 100.0 & 100.0 \\
\midrule
\multirow{11}{*}{\texttt{combined}} & bn & 3507.6 & BENGALI: 99.6, CHINESE: 0.4 & 99.6 \\
& de & 1845.2 & GERMAN: 100.0 & 100.0 \\
& en & 1582.6 & ENGLISH: 100.0 & 100.0 \\
& es & 2604.3 & SPANISH: 100.0 & 100.0 \\
& fr & 1726.5 & FRENCH: 99.2, CHINESE: 0.8 & 99.2 \\
& ja & 2127.4 & JAPANESE: 100.0 & 100.0 \\
& ru & 1523.6 & RUSSIAN: 98.8, ENGLISH: 0.4, GERMAN: 0.4, CHINESE: 0.4 & 98.8 \\
& sw & 3161.1 & SWAHILI: 98.4, ENGLISH: 1.6 & 98.4 \\
& te & 7036.0 & TELUGU: 97.6, ENGLISH: 2.4 & 97.6 \\
& th & 2043.3 & THAI: 90.8, CHINESE: 8.8, GERMAN: 0.4 & 90.8 \\
& zh & 1188.7 & CHINESE: 100.0 & 100.0 \\
\bottomrule
\end{tabular}
}

\label{tab:lang-compliance-32B}
\end{table}

\begin{table}[t]
\caption{Results for \textbf{s1-14B}.}
\centering
\resizebox{\textwidth}{!}{%
\begin{tabular}{l|l|r|l|r}
\toprule
\textbf{Strategy} & \textbf{Reasoning Language} & \textbf{avg len} & \textbf{Language Distribution} & \textbf{Language Compliance} \\ 
\midrule
\multirow{11}{*}{\texttt{translated\_wait}}     & bn & 2457.1 & ENGLISH: 100.0 & 0.0 \\
& de & 1940.1 & ENGLISH: 76.4, GERMAN: 23.6 & 23.6 \\
& en & 1638.2 & ENGLISH: 100.0 & 100.0 \\
& es & 1966.1 & ENGLISH: 94.4, SPANISH: 5.6 & 5.6 \\
& fr & 2062.6 & ENGLISH: 95.2, FRENCH: 4.4, CHINESE: 0.4 & 4.4 \\
& ja & 2162.1 & ENGLISH: 86.8, JAPANESE: 13.2 & 13.2 \\
& ru & 1937.9 & ENGLISH: 98.8, RUSSIAN: 0.8, CHINESE: 0.4 & 0.8 \\
& sw & 3044.8 & ENGLISH: 99.6, SWAHILI: 0.4 & 0.4 \\
& te & 3852.4 & ENGLISH: 98.8, CHINESE: 0.4, TELUGU: 0.8 & 0.8 \\
& th & 1767.3 & ENGLISH: 100.0 & 0.0 \\
& zh & 1438.9 & ENGLISH: 62.0, CHINESE: 38.0 & 38.0 \\
\midrule
\multirow{11}{*}{\texttt{prefix}}   & bn & 4570.1 & BENGALI: 92.8, ENGLISH: 7.2 & 92.8 \\
& de & 1621.8 & GERMAN: 100.0 & 100.0 \\
& en & 1572.2 & ENGLISH: 100.0 & 100.0 \\
& es & 2343.2 & SPANISH: 99.6, ENGLISH: 0.4 & 99.6 \\
& fr & 1354.3 & FRENCH: 99.6, ENGLISH: 0.4 & 99.6 \\
& ja & 2194.3 & JAPANESE: 99.2, ENGLISH: 0.8 & 99.2 \\
& ru & 1470.3 & RUSSIAN: 98.8, CHINESE: 0.8, ENGLISH: 0.4 & 98.8 \\
& sw & 4442.6 & SWAHILI: 85.2, ENGLISH: 14.4, TAGALOG: 0.4 & 85.2 \\
& te & 6041.8 & TELUGU: 90.4, ENGLISH: 9.2, CHINESE: 0.4 & 90.4 \\
& th & 2448.6 & THAI: 88.4, CHINESE: 10.8, ENGLISH: 0.8 & 88.4 \\
& zh & 1149.3 & CHINESE: 100.0 & 100.0 \\
\midrule
\multirow{11}{*}{\texttt{system}}   & bn & 2196.2 & ENGLISH: 99.2, BENGALI: 0.8 & 0.8 \\
& de & 1643.4 & GERMAN: 92.8, ENGLISH: 7.2 & 92.8 \\
& en & 1664.3 & ENGLISH: 100.0 & 100.0 \\
& es & 1766.7 & ENGLISH: 84.4, SPANISH: 15.6 & 15.6 \\
& fr & 1841.7 & ENGLISH: 91.6, FRENCH: 8.4 & 8.4 \\
& ja & 2036.4 & JAPANESE: 17.2, ENGLISH: 82.8 & 17.2 \\
& ru & 1631.3 & RUSSIAN: 49.2, ENGLISH: 50.8 & 49.2 \\
& sw & 2984.6 & ENGLISH: 98.8, SWAHILI: 1.2 & 1.2 \\
& te & 3942.8 & ENGLISH: 91.6, TELUGU: 8.4 & 8.4 \\
& th & 1775.4 & ENGLISH: 91.6, THAI: 8.0, CHINESE: 0.4 & 8.0 \\
& zh & 1295.8 & ENGLISH: 12.0, CHINESE: 88.0 & 88.0 \\
\midrule
\multirow{11}{*}{\texttt{combined}} & bn & 4823.5 & BENGALI: 98.0, ENGLISH: 2.0 & 98.0 \\
 & de & 1479.8 & GERMAN: 100.0 & 100.0 \\
 & en & 1553.2 & ENGLISH: 100.0 & 100.0 \\
 & es & 2503.3 & SPANISH: 100.0 & 100.0 \\
 & fr & 1269.8 & FRENCH: 100.0 & 100.0 \\
 & ja & 1941.4 & JAPANESE: 100.0 & 100.0 \\
 & ru & 1577.6 & RUSSIAN: 99.6, CHINESE: 0.4 & 99.6 \\
 & sw & 4525.5 & SWAHILI: 88.4, ENGLISH: 11.6 & 88.4 \\
 & te & 6046.9 & TELUGU: 93.6, ENGLISH: 6.4 & 93.6 \\
 & th & 2244.8 & THAI: 92.4, CHINESE: 7.2, ENGLISH: 0.4 & 92.4 \\
 & zh & 1118.8 & CHINESE: 100.0 & 100.0 \\
\bottomrule
\end{tabular}
}

\label{tab:lang-compliance-14B}
\end{table}

\begin{table}[t]
\caption{Results for \textbf{s1-7B}.}
\centering
\resizebox{\textwidth}{!}{%
\begin{tabular}{l|l|r|l|r}
\toprule
\textbf{Strategy} & \textbf{Reasoning Language} & \textbf{avg len} & \textbf{Language Distribution} & \textbf{Language Compliance} \\ 
\midrule
\multirow{11}{*}{\texttt{translated\_wait}}     & bn & 3355.8 & ENGLISH: 100.0 & 0.0 \\
& de & 2490.5 & ENGLISH: 68.0, GERMAN: 32.0 & 32.0 \\
& en & 2050.4 & ENGLISH: 100.0 & 100.0 \\
& es & 3156.8 & ENGLISH: 83.6, SPANISH: 16.4 & 16.4 \\
& fr & 2544.1 & ENGLISH: 90.4, FRENCH: 9.2, CHINESE: 0.4 & 9.2 \\
& ja & 3381.9 & ENGLISH: 72.0, JAPANESE: 28.0 & 28.0 \\
& ru & 2742.3 & ENGLISH: 96.0, RUSSIAN: 2.8, CHINESE: 1.2 & 2.8 \\
& sw & 5381.5 & ENGLISH: 100.0 & 0.0 \\
& te & 5232.8 & ENGLISH: 97.6, TELUGU: 2.4 & 2.4 \\
& th & 2654.8 & ENGLISH: 89.6, THAI: 8.4, CHINESE: 2.0 & 8.4 \\
& zh & 1471.1 & CHINESE: 99.2, ENGLISH: 0.8 & 99.2 \\
\midrule
\multirow{11}{*}{\texttt{prefix}}   & bn & 3814.5 & BENGALI: 91.6, ENGLISH: 7.2, CHINESE: 1.2 & 91.6 \\
& de & 2183.4 & GERMAN: 98.0, CHINESE: 1.6, ENGLISH: 0.4 & 98.0 \\
& en & 2405.3 & ENGLISH: 100.0 & 100.0 \\
& es & 3468.3 & SPANISH: 99.2, CHINESE: 0.4, ENGLISH: 0.4 & 99.2 \\
& fr & 1712.5 & FRENCH: 99.2, CHINESE: 0.8 & 99.2 \\
& ja & 4976.8 & JAPANESE: 99.6, ENGLISH: 0.4 & 99.6 \\
& ru & 2242.7 & RUSSIAN: 82.4, CHINESE: 13.6, ENGLISH: 4.0 & 82.4 \\
& sw & 7653.5 & SWAHILI: 55.2, ENGLISH: 43.6, TAGALOG: 1.2 & 55.2 \\
& te & 6649.8 & TELUGU: 82.4, ENGLISH: 17.2, CHINESE: 0.4 & 82.4 \\
& th & 3239.8 & THAI: 80.0, CHINESE: 17.6, ENGLISH: 2.4 & 80.0 \\
& zh & 1675.4 & CHINESE: 100.0 & 100.0 \\
\midrule
\multirow{11}{*}{\texttt{system}}   & bn & 3508.5 & ENGLISH: 96.4, BENGALI: 3.2, CHINESE: 0.4 & 3.2 \\
   & de & 2093.8 & GERMAN: 97.2, CHINESE: 2.8 & 97.2 \\
   & en & 2721.8 & ENGLISH: 100.0 & 100.0 \\
   & es & 4099.5 & SPANISH: 100.0 & 100.0 \\
   & fr & 1872.1 & FRENCH: 99.6, CHINESE: 0.4 & 99.6 \\
   & ja & 2513.9 & JAPANESE: 100.0 & 100.0 \\
   & ru & 2166.5 & RUSSIAN: 79.2, CHINESE: 9.2, ENGLISH: 11.6 & 79.2 \\
   & sw & 5310.8 & ENGLISH: 100.0 & 0.0 \\
   & te & 5250.4 & ENGLISH: 96.8, TELUGU: 3.2 & 3.2 \\
   & th & 3438.9 & THAI: 93.2, CHINESE: 6.8 & 93.2 \\
   & zh & 1528.5 & CHINESE: 100.0 & 100.0 \\
\midrule
\multirow{11}{*}{\texttt{combined}} & bn & 4094.8 & BENGALI: 97.6, CHINESE: 0.8, ENGLISH: 1.6 & 97.6 \\
 & de & 2046.7 & GERMAN: 100.0 & 100.0 \\
 & en & 2857.6 & ENGLISH: 100.0 & 100.0 \\
 & es & 3907.6 & SPANISH: 100.0 & 100.0 \\
 & fr & 1657.1 & FRENCH: 99.6, CHINESE: 0.4 & 99.6 \\
 & ja & 4804.9 & JAPANESE: 100.0 & 100.0 \\
 & ru & 2450.9 & RUSSIAN: 86.4, CHINESE: 13.2, ENGLISH: 0.4 & 86.4 \\
 & sw & 7393.2 & SWAHILI: 81.6, ENGLISH: 18.4 & 81.6 \\
 & te & 6403.0 & TELUGU: 98.4, ENGLISH: 1.6 & 98.4 \\
 & th & 3609.9 & THAI: 92.8, CHINESE: 7.2 & 92.8 \\
 & zh & 1734.0 & CHINESE: 100.0 & 100.0 \\
\bottomrule
\end{tabular}
}

\label{tab:lang-compliance-7B}
\end{table}

\subsection{In-Language Budget Forcing Reasoning}
\label{app:detailed_results_in_lang}
\begin{table}[t]
\caption{Performance comparison of different language forcing strategies across multiple model sizes and languages on fixed 8k thinking tokens. Languages are categorized into high-resource (HRL: \texttt{de}, \texttt{en}, \texttt{es}, \texttt{fr}, \texttt{ru}, \texttt{ja}, \texttt{zh}) and low-resource (LRL: \texttt{bn}, \texttt{sw}, \texttt{te}, \texttt{th}) groups. 
}
    \centering
    \small
     \scalebox{0.75}{
    \begin{tabular}{llccccccccccc|ccc}
        \toprule
        \textbf{Model} & \textbf{Method} & \textbf{bn} & \textbf{de} & \textbf{en} & \textbf{es} & \textbf{fr} & \textbf{ja} & \textbf{ru} & \textbf{sw} & \textbf{te} & \textbf{th} & \textbf{zh} & \textbf{ALL} & \textbf{HRL} & \textbf{LRL}\\
        \midrule
        \multirow{5}{*}{s1.1-32B} & \texttt{Baseline} & 90.8 & \textbf{90.8} & 96.0 & \textbf{93.2} & 89.6 & 87.6 & 93.2 & 72.4 & 68.4 & \textbf{91.6} & 88.0 & 87.4 & 91.2 & 80.8 \\
        & \texttt{translated\_wait} & \textbf{91.2} & 90.4 & 94.8 & \textbf{93.2} & 89.2 & \textbf{89.2} & 92.0 & \textbf{73.2} &\textbf{ 70.8} & \textbf{91.6} & 90.0 & \textbf{87.8} & \textbf{91.3} & \textbf{81.7}\\
        & \texttt{prefix} & 85.2 & 90.4 & 95.6 & 92.8 & \textbf{90.4} & 84.8 & \textbf{94.0} & 65.2 & 63.6 & 88.8 & \textbf{90.8} & 85.6 & \textbf{91.3} & 75.7 \\
        & \texttt{system} & 87.6 & 90.0 & \textbf{96.4} & 91.2 & 86.8 & 85.2 & 92.8 & 71.2 & 67.2 & 90.8 & 89.6 & 86.0 & 90.3 & 79.3\\
        & \texttt{combined} & 82.8 & 89.2 & 95.2 & 91.6 & 88.8 & 85.2 & 92.8 & 58.4 & 63.2 & 89.2 & 90.4 & 84.3 & 90.5 & 73.4\\
        \midrule
        \multirow{5}{*}{s1.1-14B} & \texttt{Baseline} & \textbf{86.8} & \textbf{90.4} & 94.4 & \textbf{93.6} & 88.4 & 83.6 & 92.4 & 59.6 & 60.0 & 89.2 & 89.6 & 84.4 & 90.3 & 73.9 \\
        &\texttt{translated\_wait} & 85.6 & 90.0 & \textbf{96.8} & \textbf{93.6} & 86.8 & \textbf{85.2} & 92.4 & \textbf{63.2} & 61.2 & \textbf{90.8} & 89.2 & \textbf{85.0} & 90.6 & \textbf{75.2}\\
        & \texttt{prefix} & 81.2 & \textbf{90.4} & 95.2 & 92.0 & \textbf{90.4} & 82.8 & \textbf{92.8} & 44.4 & 55.2 & 86.8 & \textbf{91.2} & 82.0 & \textbf{90.7} & 66.9\\
        & \texttt{system} & 84.0 & 88.8 & 95.2 & 91.2 & 87.2 & 82.8 & 91.2 & 58.8 & \textbf{62.0} & 90.4 & 90.4 & 83.8 & 89.5 & 73.8\\
        & \texttt{combined} & 81.2 & 90.0 & 93.2 & 92.0 & 86.4 & 82.4 & 90.4 & 36.8 & 54.4 & 89.2 & 88.4 & 80.4 & 89.0 & 65.4\\
        \midrule
        \multirow{5}{*}{s1.1-7B} & \texttt{Baseline} & \textbf{72.0} & \textbf{87.6} & 92.4 & 88.8 & 83.2 & \textbf{82.4} & 88.0 & 24.0 & 36.8 & 81.6 & \textbf{86.0} & \textbf{74.8} & \textbf{86.9} & \textbf{53.6} \\
        & \texttt{translated\_wait} & 69.2 & 84.0 & 93.2 & 89.2 & \textbf{87.2} & 76.4 & 87.2 & 24.0 & \textbf{37.2} & \textbf{84.0} & 83.2 & 74.1 & 85.8 & \textbf{53.6} \\
        & \texttt{prefix} & 64.0 & 82.8 & \textbf{93.6} & 86.8 & \textbf{87.2} & 68.0 & 85.6 & 14.4 & 24.0 & 74.4 & 84.0 & 69.5 & 84.0 & 44.2\\
        & \texttt{system} & 71.6 & 84.0 & 90.8 & \textbf{92.0} & 82.8 & 74.4 & \textbf{87.6} & \textbf{25.6} & 36.8 & 76.8 & 82.0 & 73.1 & 84.8 & 52.7\\
        & \texttt{combined} & 60.8 & 84.0 & 92.8 & 88.0 & 83.6 & 72.8 & 86.4 & 14.8 & 27.6 & 74.4 & 83.6 & 69.9 & 84.5 & 44.4\\
        \bottomrule
    \end{tabular}
    }
    
    \label{tab:lang_forcing_full}
\end{table}

Detailed results for in-language budget forcing reasoning performance on MGSM are presented in Table \ref{tab:lang_forcing_full}.

\begin{table}[t]
\caption{Number of average thinking tokens for each reasoning langauge when the MGSM task questions are asked in a particular query language.}
\label{tab:token_count_NxN}
    \centering
    \scalebox{0.85}{
        \begin{tabular}{l|ccccccccccc|c}
        \toprule
        \textbf{Query}&\multicolumn{11}{c|}{\textbf{Reasoning Language}} & \multirow{2}{*}{\textbf{AVG}}\\
         \textbf{Language}& \textbf{bn} & \textbf{de} & \textbf{en} & \textbf{es} & \textbf{fr} & \textbf{ja} & \textbf{ru} & \textbf{sw} & \textbf{te} & \textbf{th} & \textbf{zh} & \\
        \midrule
        bn & 4,559 & 1,974 & 2,190 & 3,473 & 1,640 & 2,276 & 1,977 & 4,900 & 5,461 & 2,852 & 1,710 & 3,001\\
        de & 3,838 & 1,642 & 1,910 & 2,838 & 1,159 & 1,794 & 1,617 & 3,510 & 5,073 & 2,025 & 1,327 & 2,430\\
        en & 3,429 & 1,239 & 1,467 & 2,253 & 1,075 & 1,625 & 1,341 & 3,431 & 5,154 & 1,703 & 1,075 & 2,162\\
        es & 3,736 & 1,388 & 1,779 & 2,512 & 1,225 & 1,665 & 1,467 & 3,600 & 5,323 & 1,875 & 1,133 & 2,336\\
        fr & 3,868 & 1,562 & 1,886 & 2,886 & 1,218 & 2,103 & 1,577 & 4,077 & 5,033 & 2,218 & 1,263 & 2,517\\
        ja & 3,994 & 1,623 & 2,028 & 2,627 & 1,366 & 2,033 & 1,616 & 3,957 & 5,334 & 2,285 & 1,333 & 2,563\\
        ru & 4,105 & 1,392 & 1,713 & 2,695 & 1,170 & 2,007 & 1,469 & 4,055 & 4,818 & 2,132 & 1,098 & 2,423\\
        sw & 5,958 & 2,945 & 2,955 & 3,801 & 2,144 & 4,114 & 3,108 & 3,867 & 6,836 & 3,976 & 2,705 & 3,855\\
        te & 5,761 & 3,341 & 3,528 & 4,160 & 2,473 & 3,574 & 3,337 & 5,934 & 6,277 & 3,879 & 2,557 & 4,074\\
        th & 4,166 & 1,687 & 2,000 & 2,801 & 1,300 & 1,993 & 1,756 & 4,107 & 5,247 & 2,279 & 1,241 & 2,597\\
        zh & 4,017 & 1,650 & 1,737 & 2,521 & 1,365 & 1,822 & 1,649 & 4,233 & 5,344 & 2,009 & 1,117 & 2,496\\
        \midrule
        AVG & 4,311 & 1,858 & 2,108 & 2,960 & 1,466 & 2,273 & 1,901 & 4,151 & 5,445 & 2,475 & 1,505 & {$^\searrow$}2,585\\

        \bottomrule
    \end{tabular}
    }
\end{table}

\section{Further Details on Cross-Domain Generalization}
\subsection{Evaluation Details}\label{app:mmlu-eval-details}
For GlobalMMLU, we use GPT-4o-mini to parse the final answer since their formats can vary in the reasoning traces. We set \texttt{max\_tokens} to 2 to check if the extracted answer contains any of the four options (A, B, C or D). See the prompts used for evaluation in Box~\ref{box:mmlu-eval}.
\begin{CJK}{UTF8}{min}
\begin{example}[GlobalMMLU Eval Prompt Template]
{box:mmlu-eval}

\textbf{System}: Your task is to extract the answer choice from the Response field. Do not attempt to answer the question in the Question field yourself.

\textbf{Task Prompt}: Your task is to extract the answer (A, B, C, or D) from the generated response based on the question and the option choices.
\\
\\
Question: \texttt{\{question\}} \\
Answer choices: \\
\texttt{\{answer\_options\}}
\\
\\
Response: \texttt{\{response\}}

\end{example}
\end{CJK}

\subsection{Language Forcing for GlobalMMLU} \label{app:mmlu-language-forcing}

\begin{table}[t]
\centering
\caption{Model performances comparison with and without using \texttt{combined} strategy. \textcolor{lightgreen}{Green} cells show positive gains compared to no-forcing setting and \textcolor{lightred}{red} for the opposite.}
\label{tab:mmlu_forcing}
\resizebox{0.9\textwidth}{!}{%
\begin{tabular}{@{}clllllllll@{}}
\\
\toprule
\multicolumn{1}{l}{} &      &                 & \multicolumn{7}{c}{\textbf{Languages}}                \\ \cmidrule(l){4-10} 
\multicolumn{1}{l}{\multirow{-2}{*}{\textbf{Model}}} &
  \multirow{-2}{*}{\textbf{Method}} &
  \multirow{-2}{*}{\textbf{Domain}} &
  \textbf{bn} &
  \textbf{de} &
  \textbf{en} &
  \textbf{fr} &
  \textbf{ja} &
  \textbf{sw} &
  \textbf{zh} \\ \midrule
                     & none & STEM            & 95.65 & 93.48 & 93.48 & 95.65 & 93.48 & 76.09 & 93.48 \\
                     &      & business        & 74.14 & 79.31 & 91.38 & 82.76 & 89.66 & 46.55 & 81.03 \\
                     &      & humanities      & 69.61 & 83.33 & 88.24 & 81.37 & 79.41 & 45.10 & 80.39 \\
                     &      & medical         & 66.67 & 77.78 & 80.56 & 72.22 & 69.44 & 52.78 & 77.78 \\
                     &      & other           & 76.79 & 87.50 & 87.50 & 83.93 & 85.71 & 66.07 & 80.36 \\
                     &      & social sciences & 71.57 & 82.35 & 89.22 & 83.33 & 80.39 & 47.06 & 84.31 \\
                     &      & AVG             & 75.74 & 83.96 & 88.39 & 83.21 & 83.02 & 55.61 & 82.89 \\ \cmidrule(l){2-10} 
 &
  combined &
  STEM &
  \cellcolor[HTML]{FFCCC9}89.13 &
  \cellcolor[HTML]{A8E6A9}95.65 &
  \cellcolor[HTML]{A8E6A9}97.83 &
  \cellcolor[HTML]{A8E6A9}97.83 &
  \cellcolor[HTML]{FFCCC9}86.96 &
  \cellcolor[HTML]{FFCCC9}67.39 &
  \cellcolor[HTML]{FFCCC9}91.30 \\
 &
   &
  business &
  \cellcolor[HTML]{FFCCC9}67.24 &
  \cellcolor[HTML]{A8E6A9}82.76 &
  \cellcolor[HTML]{FFCCC9}87.93 &
  \cellcolor[HTML]{FFCCC9}81.03 &
  \cellcolor[HTML]{FFCCC9}82.76 &
  \cellcolor[HTML]{FFCCC9}32.76 &
  \cellcolor[HTML]{A8E6A9}86.21 \\
 &
   &
  humanities &
  \cellcolor[HTML]{FFCCC9}67.65 &
  \cellcolor[HTML]{FFCCC9}78.43 &
  \cellcolor[HTML]{FFCCC9}87.25 &
  \cellcolor[HTML]{FFCCC9}79.41 &
  \cellcolor[HTML]{FFCCC9}74.51 &
  \cellcolor[HTML]{A8E6A9}49.02 &
  \cellcolor[HTML]{FFCCC9}79.41 \\
 &
   &
  medical &
  \cellcolor[HTML]{FFCCC9}63.89 &
  \cellcolor[HTML]{A8E6A9}83.33 &
  80.56 &
  72.22 &
  \cellcolor[HTML]{A8E6A9}77.78 &
  \cellcolor[HTML]{FFCCC9}44.44 &
  \cellcolor[HTML]{FFCCC9}75.00 \\
 &
   &
  other &
  \cellcolor[HTML]{FFCCC9}66.07 &
  \cellcolor[HTML]{FFCCC9}80.36 &
  \cellcolor[HTML]{A8E6A9}94.64 &
  \cellcolor[HTML]{A8E6A9}87.50 &
  \cellcolor[HTML]{FFCCC9}76.79 &
  \cellcolor[HTML]{FFCCC9}64.29 &
  \cellcolor[HTML]{FFCCC9}76.79 \\
 &
   &
  social sciences &
  \cellcolor[HTML]{FFCCC9}64.71 &
  \cellcolor[HTML]{A8E6A9}84.31 &
  \cellcolor[HTML]{FFCCC9}88.24 &
  \cellcolor[HTML]{FFCCC9}81.37 &
  \cellcolor[HTML]{A8E6A9}82.35 &
  \cellcolor[HTML]{FFCCC9}41.18 &
  \cellcolor[HTML]{FFCCC9}80.39 \\
\multirow{-14}{*}{s1.1-32B} &
   &
  AVG &
  \cellcolor[HTML]{FFCCC9}69.78 &
  \cellcolor[HTML]{A8E6A9}84.14 &
  \cellcolor[HTML]{A8E6A9}89.41 &
  \cellcolor[HTML]{A8E6A9}83.23 &
  \cellcolor[HTML]{FFCCC9}80.19 &
  \cellcolor[HTML]{FFCCC9}49.85 &
  \cellcolor[HTML]{FFCCC9}81.52 \\ \bottomrule
\end{tabular}%
}
\end{table}

We further study if forcing s1 to reason in the same language as the question helps with performance, as different languages may have different knowledge recall mechanism \citep{fierro2024multilingual}, and retrieving the right knowledge will help with MMLU-style questions \citep{wang2024mmlupro}. 
Following the findings in Section~\ref{sec:language-forcing}, we use the \texttt{combined} language forcing strategy that yields the highest language compliance in multilingual generations.

Table~\ref{tab:mmlu_forcing} shows marginal performance improvements for reasoning language in HRLs with Latin scripts, such as German (de), English (en) and French (fr). We want to highlight that the performance gain from language forcing is inconsistent: for instance, in-language reasoning for German \texttt{business} subject questions improves accuracy by around 3\% but decreases accuracy for \texttt{humanities} subject by nearly 5\%.
For Chinese, even though it is one of the highly represented languages during pretraining of the base model Qwen, reasoning in Chinese introduces a minor performance degradation even for STEM domain. For the even lower-resourced languages like Bengali (bn) and Swahili (sw), language forcing leads to performance degradation---there is around 6\% decrease on average across domains. In general, the patterns for non-math-specific domains are similar that reasoning in \textbf{HRLs} would be more beneficial compared to \textbf{LRLs}; nonetheless the latter could be more friendly in a user-facing setting and could potentially help with easier verification for more cultural-specific tasks.

\end{document}